%% file: main.tex
\documentclass[10pt,twocolumn,letterpaper]{article}

\usepackage[pagenumbers]{cvpr} 
\usepackage{colortbl}
\input{preamble}

\def\Pluecker{Pl\"ucker~}
\def\Groebner{Gr\"obner~}
\def\OMEGA{\boldsymbol{\omega}}
\def\line{\ell}
\def\Line{^\ell}
\newcommand\boldhat[1]{%
\texorpdfstring{\hat{\mathbf{#1}}}{}}

\definecolor{cvprblue}{rgb}{0.21,0.49,0.74}
\usepackage[pagebackref,breaklinks,colorlinks,citecolor=cvprblue]{hyperref}

\definecolor{somegray}{rgb}{0.5, 0.5, 0.5}
\newcommand{\darkgrayed}[1]{\textcolor{somegray}{#1}}
\makeatletter
\newcommand*\titleheader[1]{\gdef\@titleheader{#1}}
\AtBeginDocument{%
  \let\st@red@title\@title
  \def\@title{%
    \vskip-5em
    \bgroup\normalfont\large\centering\@titleheader\par\egroup
    \vskip1.5em\st@red@title}
}
\makeatother

\titleheader{\darkgrayed{This paper has been accepted for publication at the\\
IEEE/CVF Conference on Computer Vision and Pattern Recognition (CVPR), Seattle, 2024.
\copyright IEEE}}

\def\paperID{9447} 
\def\confName{CVPR}
\def\confYear{2024}

\title{An N-Point Linear Solver for Line and Motion Estimation with Event Cameras}

\author{Ling Gao$^{1}$\thanks{indicates equal contribution} \quad Daniel Gehrig$^{2}$\footnotemark[1] \quad Hang Su$^1$ \quad Davide Scaramuzza$^2$ \quad Laurent Kneip$^1$\\ \\
$^1$ Mobile Perception Lab, ShanghaiTech University, China \\
$^2$ Robotics and Perception Group, University of Zurich, Switzerland
}

\begin{document}
\maketitle
\input{sec/0_abstract}    
\input{sec/1_intro}
\input{sec/2_related_work}
\input{sec/3_methodology}
\input{sec/4_implementation}
\input{sec/5_experiments}
\input{sec/6_conclusion}
\input{sec/7_acknowledgments}
\input{sec/X_suppl}

{
  \small
  \bibliographystyle{ieeenat_fullname}
  \bibliography{main}
}


\end{document}

%% file: preamble.tex
\usepackage[dvipsnames]{xcolor}

\usepackage{amsmath,amsfonts}
\usepackage{algorithmic}
\usepackage{algorithm}
\usepackage{array}
\usepackage{textcomp}
\usepackage{stfloats}
\usepackage{url}
\usepackage{verbatim}
\usepackage{graphicx}
\usepackage{cite}

\usepackage{times}
\usepackage{epsfig}
\usepackage{float}
\usepackage{wrapfig}
\usepackage{amssymb,amsthm}
\usepackage{bm,xspace}
\usepackage{comment}
\usepackage{verbatim}
\usepackage{multirow}
\usepackage{balance}
\usepackage{booktabs}
\usepackage{etoolbox,siunitx}
\usepackage{calc}
\usepackage{pifont,hologo}
\usepackage{nicefrac}
\usepackage{siunitx}

\def\secref#1{Sec.~\ref{#1}}
\def\figref#1{Fig.~\ref{#1}}
\def\tabref#1{Tab.~\ref{#1}}
\def\eqref#1{Eq.~\ref{#1}}
\def\algref#1{Alg.~\ref{#1}}

\makeatletter
\DeclareRobustCommand\onedot{\futurelet\@let@token\@onedot}
\def\@onedot{\ifx\@let@token.\else.\null\fi\xspace}
\def\eg{\emph{e.g}\onedot} 
\def\ie{\emph{i.e}\onedot}

\def\etal{\emph{et~al}\onedot}
\def\etalcite#1{\etal~\cite{#1}}
\makeatother

\DeclareSIUnit\pixel{px}

%% file: sec/0_abstract.tex
\begin{abstract}
Event cameras respond primarily to edges---formed by strong gradients---and are thus particularly well-suited for line-based motion estimation. Recent work has shown that events generated by a single line each satisfy a polynomial constraint which describes a manifold in the space-time volume. Multiple such constraints can be solved simultaneously to recover the partial linear velocity and line parameters. In this work, we show that, with a suitable line parametrization, this system of constraints is actually linear in the unknowns, which allows us to design a novel \emph{linear solver}. Unlike existing solvers, our linear solver \emph{(i)} is fast and numerically stable since it does not rely on expensive root finding, \emph{(ii)} can solve both minimal and overdetermined systems with more than 5 events (\ie~$N \geq 5$), and \emph{(iii)} admits the characterization of all degenerate cases and multiple solutions. The found line parameters are singularity-free and have a fixed scale, which eliminates the need for auxiliary constraints typically encountered in previous work. To recover the full linear camera velocity we fuse observations from multiple lines with a novel velocity averaging scheme that relies on a geometrically-motivated residual, and thus solves the problem more efficiently than previous schemes which minimize an algebraic residual. Extensive experiments in synthetic and real-world settings demonstrate that our method surpasses the previous work in numerical stability, and operates over 600 times faster.
\end{abstract}

\noindent \textbf{Project page: \ \footnotesize\url{https://mgaoling.github.io/eventail/}}

%% file: sec/1_intro.tex
\section{Introduction}
\label{sec:intro}

Man-made scenes contain a multitude of straight lines, and exploiting these lines for motion estimation is an important feature of modern mobile vision systems like AR/VR devices and robotic systems~\cite{huang2020lidar,lu2021line,yousif2015overview,lim2022uv}. However, computer vision algorithms aimed at leveraging these line features still suffer from fundamental limitations when using standard frame-based sensing: During high-speed motion and challenging illumination conditions, these sensors suffer from motion blur and saturation effects, which have deleterious effects on line feature extraction. Event cameras~\cite{gallego2020event} are biologically inspired sensors that address the limitation of frame-based sensors by instead only measuring the \emph{changes in intensity} at a per-pixel level, and they do this with high dynamic range, low motion blur, high temporal resolution, and high spatial data-sparsity.

Due to their working principle, event cameras respond primarily to edges---formed by strong gradients---and are thus particularly well-suited for line-based motion estimation. A recent breakthrough~\cite{gao2023eventail} in event-based motion estimation introduced an incidence relation that enforces the intersection of bearing vectors emitted by events and a corresponding line that generates those events. Using this relation, a 5-point minimal solver was designed that recovers the parameters of a minimal two-point-two-plane parametrization of the line and two velocity ratios in the plane perpendicular to the line, using the \Groebner basis method and polynomial elimination theory. However, this solver suffers from several limitations: First, it relies on a non-minimal line representation using four degrees of freedom (DoF) that \emph{(i)} fails to realize that, in the absence of scale, only three DoF are needed, and \emph{(ii)} encounters singularities when describing lines parallel to the two planes. Secondly, to solve for the motion and line parameters, previous work employs a polynomial solver, which \emph{(i)} is by definition minimal and thus incapable of incorporating more than five events, and \emph{(ii)} relies on root finding algorithms that are expensive to run and suffer from instabilities that are not easily detected.

This work addresses these limitations with two important innovations: First, it introduces a new line representation based on the angle-axis representation of a rotation matrix, which is singularity-free and only depends on three DoF and thus implicitly enforces scale ambiguity. Second, in formulating the incidence relation with this new parametrization, we derive a simple algorithm for determining motion and line parameters that only relies on solving a linear system and simple vector operations and is thus orders of magnitude faster than the polynomial solver in~\cite{gao2023eventail}. This linear system is easily extended to $N > 5$ events, with a minimal increase in complexity, which enables solution refinement with inliers when employing random sample consensus (RANSAC) schemes.

The proposed solver sheds light on all possible cases of degenerate solutions, how they arise, and all additional solutions that arise from symmetries in the incidence relation. Finally, expressing the incidence relation in terms of our new line parametrization allows us to fully characterize and visualize the types of manifolds circumscribed by events generated by a single line. Our contributions are:
\begin{itemize}
  \item A minimal, three DoF representation of lines based on the angle-axis representation of a rotation matrix. This representation encodes a reference frame centered at the 3D line, and enforces a unit distance to the closest point. 
  \item A linear algorithm for determining line and motion parameters from a set of events triggered by a line. This algorithm is fast, and extensible to multiple events, and sheds light on degenerate and multiplicitous solutions.
  \item A simpler and faster scheme for fusing partial linear velocity measurements from multiple lines, based on geometrically-motivated constraints, instead of solving algebraic equations.
\end{itemize}
We validate our method extensively in simulated and real-world settings. In particular, our solver is on average 600 times faster, taking only \SI{3.25}{\us} for five events, compared to \SI{2046}{\us} for~\cite{gao2023eventail}, while achieving a similar performance.

%% file: sec/2_related_work.tex
\section{Related Work}
\label{sec:related_work}

Ego-motion estimation is crucial for intelligent mobile devices and thus has been the subject of extensive research over the past few decades. The sub-class of vision-based solutions comprises single-camera, stereo-camera, and multi-camera solutions that are potentially supported by an inertial measurement unit. A review would go beyond the scope of the present paper, and the reader is kindly referred to recent reviews such as the one by Cadena~\etalcite{cadena2016past}. This paper focuses on motion estimation with event cameras.

The latter is a challenging problem that is initially often addressed for constrained scenarios such as 2D motion~\cite{weikersdorfer2013simultaneous}, known depth or 3D structure~\cite{weikersdorfer2014event,censi2014low,mueggler2014event,gallego2016event,bryner2019event,chamorro2020high,zuo2022devo}, pure rotation~\cite{gallego2018unifying}, and homographic warping~\cite{gallego2019focus,kueng2016low,stoffregen2019event,liu2020globally,peng2022globally}. The community has furthermore explored the combination with other sensors such as standard cameras~\cite{kueng2016low,vidal2018ultimate,hidalgo2022event}, inertial units~\cite{vidal2018ultimate,zhu2017event}, or a second event camera~\cite{zhou2021event}. In turn, the present paper considers motion in 3D with a single event camera and in arbitrary environments. 
Different optimization-based~\cite{rebecq2016evo,le2020idol,mueggler2018continuous}, filter-based~\cite{kim2016real,zhu2017event} and learning-based~\cite{maqueda2018event,gehrig2020event} solutions have already been presented. Of particular interest to the present work are methods that rely on line features~\cite{yuan2016fast,le2020idol}.

In the spirit of original works on monocular visual odometry~\cite{nister2004visual}, the present work addresses the relatively unexplored topic of geometric incidence relationships for local relative motion calculation with an event camera. Geometric solutions remain important to date owing to their ability to find solutions with optimality guarantees, and potentially certificates, under known assumptions, unlike optimization-, filter- or learning-based solvers, which often lack these guarantees. Given that events are primarily triggered by high-gradient appearance boundaries, the dominant feature for event-based motion estimation is given by lines. Weng~\etalcite{weng1992motion} and Hartley~\etalcite{hartley1997lines} have proposed closed-form solutions for frame-based cameras. Tri-focal tensor geometry inspired the first closed-form solution for event cameras~\cite{peng2021continuous}. An important characteristic of this solver is that it relies on a local constant velocity motion model and makes use of the first-order dynamics of the camera. This is important as it permits the inclusion of the time-stamped, asynchronous measurements produced by an event camera. Nonetheless, the method by Peng~\etalcite{peng2021continuous} is not general as it still depends on approximate event-based line-feature extractors~\cite{brandli2016elised,valeiras2019event,vakhitov2019learnable} rather than events, only.

The most related works to ours are the works by Ieng~\etalcite{ieng2017event}, Seok and Lim~\cite{seok2020robust}, and Gao~\etalcite{gao2023eventail}, who propose model-based fitting of the manifold locations of the events generated by the observation of a line under motion. In particular, Gao~\etalcite{gao2023eventail} introduces an exact incidence relation that all such events must obey under constant linear velocity. It depends on observable camera motion and 3D line parameters, and thus serves as a basis for joint manifold fitting and motion estimation. Based on this foundation, the present work develops the first linear $N$-point solution to this problem, which not only unlocks unprecedented efficiency, but also a simplified understanding of degenerate geometric conditions.

%% file: sec/3_methodology.tex
\section{Methodology}
\label{sec:methodology}

Assume a calibrated event camera undergoing an arbitrary six DoF motion, while observing a set of $M$ lines $\{\mathbf{L}_i\}_{i=1}^{M}$. Each line generates a set of $N_i$ events $\mathcal{E}_i=\{e_{ij}\}_{j=1}^{N_i}$ where each event $e_{ij} = (\mathbf{x}_{ij}, t_{ij}, p_{ij})$ is characterized by its pixel coordinate $\mathbf{x}_{ij}$ in the image plane, timestamp $t_{ij}$ (with \SI{}{\us} resolution), and polarity $p_{ij}$.

For a small time window $[t_s-\Delta t, t_s + \Delta t]$, centered at reference time $t_s$, such that the camera motion can be approximated by linear dynamics, the events generated by a single line circumscribe a manifold termed \emph{eventail}~\cite{gao2023eventail}. In~\secref{sec:parametrization}, we describe the geometric incidence relation~\cite{gao2023eventail} which needs to be satisfied by events on this manifold and depends on the observable components of the velocity $\mathbf{v}$ and the line $\mathbf{L}_i$ for which we introduce a minimal parametrization. Then, in~\secref{sec:minimal_form} we will introduce a solver that recovers the line parameters, and observable linear velocity parameters from a set of $N_i \geq 5$ events that lie on the manifold. Finally, in~\secref{sec:averaging}, we will explore how to recover the full linear velocity from a set of $M \geq 2$ partial velocity observations. 


\subsection{Incidence Relationship}
\label{sec:parametrization}

We reiterate here briefly the incidence relation introduced in~\cite{gao2023eventail} and use~\figref{fig:geometry} for illustration purposes. For simplicity, we will first consider the case of one line, and thus drop the index $i$ from the variables. We furthermore express all quantities in the camera frame centered at time $t_s$. The incidence relation enforces that events are triggered by points on the line, such that the line $\mathbf{L}_{j}=[\mathbf{d}_j^\intercal\,\mathbf{m}_j^\intercal]^\intercal$ (in \Pluecker coordinates) emanating from an individual event $e_{j}$ triggered at time $t_{j}$ (orange line in~\figref{fig:geometry}) intersects the line $\mathbf{L}=[\mathbf{d}^\intercal\,\mathbf{m}^\intercal]^\intercal$ (blue line in~\figref{fig:geometry}). The condition for intersection of two non-parallel lines is
\begin{equation}
  \mathbf{d}^\intercal \mathbf{m}_j + \mathbf{m}^\intercal \mathbf{d}_j = 0 \,.
\end{equation}
\Pluecker line coordinates comprise the line direction $\mathbf{d}\in\mathbb{R}^3$ and moment $\mathbf{m}=\mathbf{P} \times \mathbf{d}$, with $\mathbf{P}\in\mathbb{R}^3$ being an arbitrary point on the line. As~\cite{gao2023eventail} for the event ray $j$ we use the camera position $\mathbf{C}[t_j]=\mathbf{P}_j$ at time $t_j$, and the event bearing vector $\mathbf{f}_{j}'=\mathbf{d}_j$ rotated into the reference frame at time $t_s$. Under first order dynamics, these are $\mathbf{C}[t_j] = t_{j}' \mathbf{v}$ and $\mathbf{f}_{j}' = \mathbf{R}[t_j] \mathbf{f}_j$ with $t_{j}' = t_j - t_s$. As~\cite{gao2023eventail}, we assume $\mathbf{R}[t_j]$ given and computed by integrating angular rate measurements $\OMEGA$ from an available inertial measurement unit (IMU) with $\mathbf{R}[t_j]=\text{exp}([\OMEGA]_\times t_{j}')$. Here $\text{exp}(\cdot)$ denotes the matrix exponential and $[\OMEGA]_\times$ denotes the skew-symmetric matrix associated with the angular rate $\OMEGA\in\mathbb{R}^3$. The \Pluecker coordinates are thus $\mathbf{L}_j=[\mathbf{f}_{j}'^\intercal\quad(\mathbf{C}[t_j] \times \mathbf{f}_{j}')^\intercal]^\intercal$ and the incidence relation becomes 
\begin{equation}
  \label{eq:nonminimal_incidence}
  \mathbf{d}^\intercal \left(\mathbf{C}\left[t_{j}\right] \times \mathbf{f}'_{j}\right) + \mathbf{m}^\intercal \mathbf{f}'_{j} = 0.
\end{equation}
The above equation relates measurements $\mathbf{f}_j$ and $t_j$ to our unknowns $\mathbf{d}, \mathbf{m}$ and $\mathbf{v}$, but is still not in a minimal form. This is because, firstly \Pluecker coordinates are not minimal, and indeed true minimal line representations only have four DoF. Second, in a monocular setup, there is scale ambiguity, and this dictates that the absolute scale of $\mathbf{v}$ and $\mathbf{L}$ is unobservable. Finally, the velocity component along the line direction is unobservable due to the aperture problem. To encapsulate these constraints, in the next part, we will introduce a line representation based on the angle-axis representation of rotation matrices. This rotation matrix simultaneously spans a coordinate frame in which we will express our camera velocity, and the aperture problem can be enforced succinctly.

\input{figures/fig_geometry.tex}


\subsection{Transition into a Minimal Form}
\label{sec:minimal_form}

The transition into minimal form consists of two steps. We first derive the angle-axis-based line representation by successively eliminating internal constraints in the \Pluecker line coordinate representation, and then proceed with finding the minimal camera velocity representation. We start by observing that scaling both $\mathbf{d}$ and $\mathbf{m}$ yields the same line, and thus we may choose to fix the scale of $\mathbf{d}$ to be unity. Next, we observe that $\mathbf{m} = \mathbf{P} \times \mathbf{d}$ is by definition perpendicular to $\mathbf{d}$. Moreover, we see that any $\mathbf{P}$ on the same line results in the same moment, and thus we may choose to select it closest to the origin, such that it is perpendicular to $\mathbf{d}$. Since the scale is unobservable, we may furthermore fix the distance from $\mathbf{P}$ to the origin to be unity. Summarizing these observations, we conclude that we may select $\mathbf{P}$ and $\mathbf{d}$ to be perpendicular unit vectors, and in particular we will select $\mathbf{P} = -\mathbf{e}^{\ell}_{3}$ and $\mathbf{d} = \mathbf{e}^{\ell}_{1}$, such that the resulting moment is $\mathbf{m} = -\mathbf{e}^{\ell}_{2}$. We visualize the three unit vectors in~\figref{fig:geometry}, and observe that they span a \emph{line-dependent} coordinate frame via the rotation matrix $\mathbf{R}_{\ell} = [\mathbf{e}^{\ell}_{1}\,\mathbf{e}^{\ell}_{2}\,\mathbf{e}^{\ell}_{3}]$. Rotation matrices belong to $\mathit{SO}(3)$, and thus it becomes apparent that this line representation can be further compressed via the matrix logarithm $\boldsymbol{\theta}_\ell = (\text{log}\left(\mathbf{R}_\ell\right))^{\vee}=[\theta^\ell_{x}\,\theta^\ell_{y}\,\theta^\ell_{z}]$, to yield only three DoF, which is a minimal line representation in the absence of scale. Here $(\cdot)^{\vee}$ maps the skew-symmetric matrix in the argument to the associated vector. 

Note that compared to the two-point-two-plane parametrization~\cite{hartley2003multiple} in~\cite{gao2023eventail}, this representation is \emph{(i)} minimal, relying on three instead of four DoF, and \emph{(ii)} more flexible, since it does need a reparametrization step for lines that are almost parallel to the $yz$-plane.

We now address the camera velocity parametrization, which we express in the line-dependent coordinate frame.
\begin{equation}
  \mathbf{v} = \mathbf{R}_{\ell} \mathbf{u}_{\ell} \,,
\end{equation}
introduces the camera velocity $\mathbf{u}_\ell=[u_{x}^{\ell}\,u_{y}^{\ell}\,u_{z}^{\ell}]$ expressed in the line coordinate frame. Using these new parametrizations, the incidence relation~\eqref{eq:nonminimal_incidence} becomes
\begin{equation}
  \label{eq:minimal_incidence}
  t_{j}'{\mathbf{e}^{\ell}_{1}}^\intercal ((\mathbf{R}_{\ell}\mathbf{u}_{\ell}) \times \mathbf{f}_{j}') - {\mathbf{f}_{j}'}^\intercal \mathbf{e}^{\ell}_2 = 0 \,.
\end{equation}
Cycling through the triple product in the first summand, \ie~$\mathbf{a}^\intercal (\mathbf{b}\times \mathbf{c}) = \mathbf{c}^\intercal (\mathbf{a}\times \mathbf{b})$, we arrive at 
\begin{equation}
  t_{j}'{\mathbf{f}_{j}'}^\intercal (\mathbf{e}^{\ell}_{1} \times (\mathbf{R}_{\ell} \mathbf{u}_{\ell})) - {\mathbf{f}_{j}'}^\intercal \mathbf{e}^{\ell}_2 = 0 \,,
\end{equation}
which can be expanded and further simplified to 
\begin{equation}
  \label{eq:incidence_final}
  t_{j}'{\mathbf{f}_{j}'}^\intercal (u^{\ell}_{z} \mathbf{e}^{\ell}_{2} - u_{y}^{\ell} \mathbf{e}^{\ell}_{3}) + {\mathbf{f}_{j}'}^\intercal \mathbf{e}^{\ell}_{2} = 0 \,.
\end{equation}
Note that due to the cross product with $\mathbf{e}^{\ell}_{1}$, the camera velocity $u^{\ell}_{x}$ becomes unobservable within this incidence relation, \ie~changing it does not affect the residual. This confirms our intuition that the aperture problem should make velocities along the line direction unobservable. As a result, we focus on only solving for $u^{\ell}_{y}$ and $u^{\ell}_{z}$. Furthermore, observe that these equations are in terms of $\mathbf{e}^{\ell}_2$ and $\mathbf{e}^{\ell}_{3}$ instead of the minimal parameters $\boldsymbol{\theta}_{\ell}$. In what follows we design the solver around recovering $\mathbf{e}^{\ell}_{2}$ and $\mathbf{e}^{\ell}_{3}$ since it yields a simpler algorithm, yet it should be remembered that the minimal representation can always be recovered using the matrix logarithm of $\mathbf{R}_{\ell}$. We now discuss how to recover the unknowns from a set of incidence relations from multiple events, which is summarized in~\algref{alg:linear_solver}.


\subsection{Five-point Minimal Solver}
\label{sec:theory}

The incidence relationship in~\eqref{eq:incidence_final} has five unknowns, three from $\boldsymbol{\theta}_\ell=[\theta^{\ell}_{x}\,\theta^{\ell}_{y}\,\theta^{\ell}_{z}]$ and two from $\mathbf{u}_{\ell} = [0\,u^{\ell}_{y}\,u^{\ell}_{z}]$, and thus can be solved by stacking a minimum of five such constraints. Since each such constraint originates from a single event, this means that five events are the minimum number to solve this system. This stack of equations is 
\begin{align}
  \nonumber t_{1}'{\mathbf{f}_{1}'}^\intercal (u^{\ell}_{z} \mathbf{e}^{\ell}_{2}-u^{\ell}_{y}\mathbf{e}^{\ell}_{3}) + {\mathbf{f}_{1}'}^\intercal \mathbf{e}^{\ell}_{2} &= 0 \\
  \nonumber t_{2}'{\mathbf{f}_{2}'}^\intercal (u^{\ell}_{z} \mathbf{e}^{\ell}_{2}-u^{\ell}_{y}\mathbf{e}^{\ell}_{3}) + {\mathbf{f}_{2}'}^\intercal \mathbf{e}^{\ell}_{2} &= 0 \\
  \nonumber \vdots & \\
  \nonumber t_{5}'{\mathbf{f}_{5}'}^\intercal (u^{\ell}_{z} \mathbf{e}^{\ell}_{2}-u^{\ell}_{y}\mathbf{e}^{\ell}_{3}) + {\mathbf{f}_{5}'}^\intercal \mathbf{e}^{\ell}_{2} &= 0 \,. 
\end{align}
This system of equations is linear in the unknowns and can be rewritten as a single matrix equation 
\begin{align}
  \label{eq:linear_system}
  \underbrace{\begin{bmatrix}
    t_{1}'{\mathbf{f}_{1}'}^\intercal & {\mathbf{f}_{1}'}^\intercal \\
    \vdots & \vdots \\
    t_{5}'{\mathbf{f}_{5}'}^\intercal & {\mathbf{f}_{5}'}^\intercal \\
  \end{bmatrix}}_{\doteq \mathbf{A} \in \mathbb{R}^{5 \times 6}}
  \underbrace{\begin{bmatrix}
    u^{\ell}_{z} \mathbf{e}^{\ell}_{2} - u^{\ell}_{y}\mathbf{e}^{\ell}_{3} \\
    \mathbf{e}^{\ell}_{2}
  \end{bmatrix}}_{\doteq \mathbf{x} \in \mathbb{R}^{6 \times 1}} = \mathbf{0} \,.
\end{align}
Note that this formulation successfully groups the terms from the events in $\mathbf{A}$ and unknowns in $\mathbf{x}$. We can thus solve for $\mathbf{x}$, and then reconstruct the unknowns from the found solution. Solving~\eqref{eq:linear_system} can be done with a singular value decomposition of $\textbf{A}$ and then selecting the last column of $\mathbf{V}$ corresponding to the smallest singular value of $\mathbf{A}$. Let us denote this solution with $\hat{\mathbf{x}}$. Note that $\hat{\mathbf{x}}$ is normalized, and, due to the homogeneous nature of~\eqref{eq:linear_system}, only known up to parity, \ie~both $\pm \hat{\mathbf{x}}$ are solutions. Note also that this procedure is not limited to using only five events, but can be applied to $N \geq 5$, however in this case the solution will no longer be exact, and instead approximate but with a globally minimal squared residual equal to the smallest singular value of $\mathbf{A}$. The ability to process more than five events sets this method apart from the solver in~\cite{gao2023eventail} which uses a fixed elimination template tailored to only five events. Next, we discuss how to recover the unknowns from a solution $\mathbf{\hat{x}}$. 


\subsection{Recovering the Unknowns from $\boldhat{x}$}
\label{sec:solution}

For simplicity, we will only treat the case with $+\hat{\mathbf{x}}$, but will state that the ambiguity in the parity of the solution to~\eqref{eq:linear_system} gives rise to the solution pairs $S_0, S_1$, and $S_2, S_3$. Remembering the definition of $\mathbf{x}$ we write that
\begin{equation}
  \label{eq:def_scale}
  \hat{\mathbf{x}} = 
  \begin{bmatrix}
    \lambda \hat{\mathbf{x}}_{1:3} \\
    \lambda \hat{\mathbf{x}}_{4:6}
  \end{bmatrix} 
  = 
  \begin{bmatrix}
    u^{\ell}_{z} \mathbf{e}^{\ell}_{2} - u^{\ell}_{y}\mathbf{e}^{\ell}_{3} \\
    \mathbf{e}^{\ell}_{2}
  \end{bmatrix} \,.
\end{equation}
Here $\lambda$ is an unknown scaling factor. However, since the last three entries of $\hat{\mathbf{x}}$ correspond to the unit vector $\mathbf{e}^{\ell}_{2}$, we can simply normalize $\hat{\mathbf{x}}$ by the length of $\hat{\mathbf{x}}_{4:6}$. In what follows we assume that this normalization is done beforehand, and thus ignore this scaling factor by setting $\lambda = 1$. Straightforward manipulation yields 
\begin{subequations}
  \label{eq:solver_1}
  \begin{align}
    \mathbf{e}^\ell_2 &= \hat{\mathbf{x}}_{4:6} \\
    u^{\ell}_{z} &= \hat{\mathbf{x}}_{1:3}^\intercal \hat{\mathbf{x}}_{4:6} \\
    u^{\ell}_{y} \mathbf{e}^{\ell}_{1} &= \hat{\mathbf{x}}_{1:3} \times \hat{\mathbf{x}}_{4:6} \,.
  \end{align}
\end{subequations}
From the last equation we can recover $u^{\ell}_{y}$ and $\mathbf{e}^{\ell}_{1}$ by taking the norm, and normalized vector
\begin{align}
  \label{eq:decomp_e1}
  u^{\ell}_{y} = \Vert \hat{\mathbf{x}}_{1:3} \times \hat{\mathbf{x}}_{4:6}\Vert \text{ and }
  \mathbf{e}^{\ell}_{1} = \frac{\hat{\mathbf{x}}_{1:3} \times \hat{\mathbf{x}}_{4:6}}{\Vert \hat{\mathbf{x}}_{1:3} \times \hat{\mathbf{x}}_{4:6}\Vert} \,.
\end{align}
Note that this decomposition is not unique, as we may simultaneously flip the signs of $\mathbf{e}^{\ell}_{1}$ and $u^{\ell}_{y}$, resulting in the same product. Finally, we can recover $\mathbf{e}^{\ell}_{3} = \mathbf{e}^{\ell}_{1} \times \mathbf{e}^{\ell}_{2}$. While $\hat{\mathbf{x}}$ is the globally optimal solution to the incidence relation constraints in~\eqref{eq:linear_system}, it is not immediately clear that the recovered $\mathbf{R}_l,\mathbf{u}_l$ are also globally optimal with respect to this constraint. As proved in the supplementary material it turns out that $\mathbf{R}_l,\mathbf{u}_l$ are globally optimal. Next, we will comment on verifying the correctness of the solution.


\subsection{Degenerate Solutions and Solution Multiplicity}
\label{sec:degeneracy}

First, we state a theorem that the only degenerate cases arise when the matrix $\mathbf{A}$ in~\eqref{eq:linear_system} is rank-deficient. This means that, if $\text{rank}(\mathbf{A})\geq 5$, the previously discussed decomposition always succeeds and yields four solutions. \\

Theorem 1: \noindent\emph{If $\text{rank}(\mathbf{A})\geq 5$, with $\mathbf{A}$ defined in~\eqref{eq:linear_system}, the decomposition of $\hat{\mathbf{x}}$ into $\mathbf{u}_{\ell}$ and $\mathbf{R}_{\ell}$ always succeeds and yields four distinct solutions. If $\text{rank}(\mathbf{A}) < 5$ the solver returns infinitely many solutions.} \\

Note that this theorem also handles cases in which the line passes through the origin. Both the proof of the above theorem and the handling of this case are described in the supplementary material. Rank deficiency of $\mathbf{A}$ occurs when events share the same timestamp $t_j$ or motion-corrected bearing vector $\mathbf{f}_{j}'$. To identify this scenario, checking the matrix rank before solving for $\hat{\mathbf{x}}$ can be done. After SVD, the second smallest singular value should be checked for being large, since small values indicate near-rank deficiency.

Next, we discuss the multiplicity of solutions. As previously stated, the designed solver returns four distinct solutions if the rank of $\mathbf{A}$ is at least 5. Here we enumerate these solutions (visualized in~\figref{fig:solution_multiplicity}), stated as a theorem: \\

Theorem 2: \emph{Given a solution $S_0 = \{\mathbf{e}^{\ell}_{1}, \mathbf{e}^{\ell}_{2}, \mathbf{e}^{\ell}_{3}, u^{\ell}_{y}, u^{\ell}_{z}\}$ to the incidence relation in~\eqref{eq:incidence_final}, then 
\begin{align}
  \nonumber S_1 &= \{ \mathbf{e}^{\ell}_{1}, -\mathbf{e}^{\ell}_{2}, -\mathbf{e}^{\ell}_{3},  u^{\ell}_{y}, u^{\ell}_{z}\} \,, \\
  \nonumber S_2 &= \{-\mathbf{e}^{\ell}_{1},  \mathbf{e}^{\ell}_{2}, -\mathbf{e}^{\ell}_{3}, -u^{\ell}_{y}, u^{\ell}_{z}\} \,, \\
  \nonumber S_3 &= \{-\mathbf{e}^{\ell}_{1}, -\mathbf{e}^{\ell}_{2},  \mathbf{e}^{\ell}_{3}, -u^{\ell}_{y}, u^{\ell}_{z}\}
\end{align}
are also solutions. For solutions $S_1$ and $S_2$ the closest point $-\mathbf{e}_3^\ell$ on the line is behind the camera, while for solutions $S_2$ and $S_3$ the line direction $\mathbf{e}^{\ell}_{1}$ is flipped, which represents an ambiguity in the definition of direction of $\mathbf{L}$.} \\

We state the proof in the supplementary material. Note that two configurations correspond to flipping across the xy-plane. We eliminate these solutions by enforcing that the intersection point $\mathbf{P}_j$ (see~\figref{fig:geometry}) between the line and event ray is in front of the camera. In the supplementary material, we use this geometric interpretation to characterize manifolds spanned by events in more detail.

\input{figures/fig_solution_multiplicity}

\begin{algorithm}[t!]
  \caption{Linear Solver for Line and Partial Motion Parameters}
  \textbf{Input:} A set of events $\mathcal{E}$ with rotated bearing vectors.\\
  \textbf{Output:} Line parameters $\boldsymbol{\theta}_{\ell}$ and projected velocities $\mathbf{u}_{\ell}$.
  \begin{itemize}
    \item Form matrix $\mathbf{A}$ from the set of events $\mathcal{E}$ by~\eqref{eq:linear_system} and make sure that $\text{rank}(\mathbf{A})\geq 5$.
    \item Apply SVD on $\mathbf{A}$ and select the last column of $\mathbf{V}$, denoted with $\hat{\mathbf{x}}$. Both $\pm\hat{\mathbf{x}}$ can be selected.  
    \item Normalize $\hat{\mathbf{x}}$ by $\hat{\mathbf{x}}_{4:6}$, the last three elements.
    \item Recover $\mathbf{e}^{\ell}_{2}, u^{\ell}_{z}$ from~\eqref{eq:solver_1}.
    \item Recover $\mathbf{e}^{\ell}_{1}, u^{\ell}_{y}$ from~\eqref{eq:decomp_e1}. Both $\{\mathbf{e}^{\ell}_{1}, u^{\ell}_{y}\}$ and $\{-\mathbf{e}^{\ell}_{1}, -u^{\ell}_{y}\}$ can be selected.
    \item Compile $\mathbf{u}_\ell = [0\,u^{\ell}_{y}\,u^{\ell}_{z}]$.
    \item Compute $\mathbf{e}^{\ell}_{3} = \mathbf{e}^{\ell}_{1} \times \mathbf{e}^{\ell}_{2}$.
    \item Construct the rotation $\mathbf{R}_\ell=[\mathbf{e}^{\ell}_{1}\,\mathbf{e}^{\ell}_{2}\,\mathbf{e}^{\ell}_{3}]$.
    \item Recover minimal line parameters $\boldsymbol{\theta}_\ell = (\text{log}\left(\mathbf{R}_\ell\right))^{\vee}$.
  \end{itemize} 
  \label{alg:linear_solver}
\end{algorithm}


\subsection{Velocity Averaging from Multiple Manifolds}
\label{sec:averaging}

As previously discussed, the linear solver can only recover partial velocities perpendicular to the line generating the events in a single cluster. We now describe how to recover the full velocity from a set of partial observations from multiple lines $\mathbf{L}_i$. Each partial observation of the velocity $\mathbf{v}$ is the projection $\hat{\mathbf{v}}_{i} = \mathbf{H}_i \mathbf{v}$ into the $\mathbf{e}^{\ell}_{2i} \mathbf{e}^{\ell}_{3i}$-plane where $\mathbf{H}_i = \mathbf{I} - \mathbf{e}^{\ell}_{1i} {\mathbf{e}^{\ell}_{1i}}^\intercal$, which is visualized in~\figref{fig:velocity_averaging}(i). The observation is given by the two projections $u^{\ell}_{yi}$ and $u^{\ell}_{zi}$ scaling the second and third basis vectors of $\mathbf{R}_{{\ell}i}$, respectively. Thus, the correct velocity estimate must satisfy
\begin{equation}
  \label{eq:line_constraint}
  \hat{\mathbf{v}}_{i}=\mathbf{H}_i\mathbf{v} = \mathbf{e}^{\ell}_{2i} u\Line_{yi} + \mathbf{e}^{\ell}_{3i} u^{\ell}_{zi} =\mathbf{R}_{{\ell}i} \mathbf{u}_{{\ell}i} \,.
\end{equation}

Unlike the velocity averaging scheme in~\cite{gao2023eventail} we adopt the following geometrically motivated, but equivalent formalism to solve multiple such equations: Each such constraint (one for each line) can be converted into a homogeneous linear constraint following the steps in~\figref{fig:velocity_averaging}(i). We see that the 90$^\circ$ rotated velocity $\mathbf{R}_{{\ell}i} \mathbf{R}_{\frac{\pi}{2}} \mathbf{u}_{{\ell}i}$ must be perpendicular to the projected camera velocity $\hat{\mathbf{v}}_{i}$. Forming the dot product with this vector yields: 
\begin{align}
  \hat{\mathbf{v}}_{i}^\intercal \mathbf{R}_{{\ell}i} \mathbf{R}_{\frac{\pi}{2}} \mathbf{u}_{{\ell}i} &=\mathbf{v}^\intercal \left(u^{\ell}_{yi} \mathbf{e}^{\ell}_{3i} - u^{\ell}_{zi} \mathbf{e}^{\ell}_{2i}\right)= 0 \\
  \nonumber\text{ with } \mathbf{R}_{\frac{\pi}{2}} &= 
  \begin{bmatrix}
    1 & 0 &  0 \\
    0 & 0 & -1 \\
    0 & 1 &  0
  \end{bmatrix} \,.
\end{align}
Note that the above constraint remains, even when inserting the alternative solutions $S_0, S_3$ or $S_1, S_2$, since these only lead to a change in the sign. Stacking $M$ such equations, one for each line, can be summarized as: 
\begin{equation}
  \label{eq:velocity_averaging}
  \underbrace{\begin{bmatrix}
    u^{\ell}_{y1} {\mathbf{e}_{31}^{\ell}}^\intercal - u^{\ell}_{z1} {\mathbf{e}_{21}^{\ell}}^\intercal \\
    \vdots \\
    u^{\ell}_{yM} {\mathbf{e}_{3M}^{\ell}}^\intercal - u^{\ell}_{zM} {\mathbf{e}_{2M}^{\ell}}^\intercal \\        
  \end{bmatrix}}_{\doteq \mathbf{D} \in \mathbb{R}^{M\times 3}} \mathbf{v} = \mathbf{0} \,.
\end{equation}

We solve this equation again with SVD, by selecting the column of $\mathbf{V}$ corresponding with the minimal singular value. From this system, we also conclude that we need a minimum of two lines to recover $\mathbf{v}$. Again, SVD only recovers $\mathbf{v}$ up to an unknown parity, \ie~both $\pm \mathbf{v}$ satisfy the equation. To disambiguate the solution, we enforce that the projected velocity must point in the same direction as $\mathbf{R}_{{\ell}i} \mathbf{u}_{{\ell}i}$ for each line. This can be written concisely as 
\begin{equation}
  \label{eq:solution_checking}
  \hat{\mathbf{v}}_{i}^\intercal \mathbf{R}_{{\ell}i} \mathbf{u}_{{\ell}i}=\mathbf{v}^\intercal \left(u^{\ell}_{yi} \mathbf{e}^{\ell}_{2i} + u^{\ell}_{zi} \mathbf{e}^{\ell}_{3i}\right) > 0 \,,
\end{equation}

If at least one line does not satisfy this condition, the sign of $\mathbf{v}$ should be flipped. Compared to~\cite{gao2023eventail}, which uses a more expensive Shur-Complement step to eliminate unknown scale factors, our algorithm only requires a single step of SVD, but yields the same results. This is because our algorithm is actually equivalent to that of~\cite{gao2023eventail}, as will be shown in the supplementary material.

Note that the above line averaging scheme may run into issues, if two lines are parallel. Making, the common $\mathbf{e}_1^\ell$ component of $\mathbf{v}$ unobservable. However, this case actually induces a rank deficiency on $\mathbf{D}$, as proved in the supplementary material, and can thus easily be discarded.

\input{figures/fig_velocity_averaging}

%% file: figures/fig_geometry.tex
\begin{figure}
  \centering
  \includegraphics[width=0.8\linewidth]{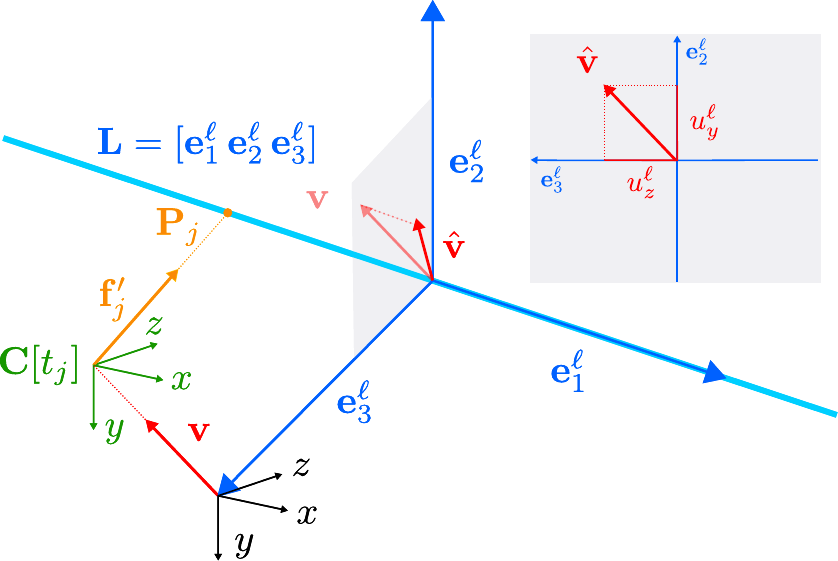}
  \caption{Incidence relationship between the line $\mathbf{L}$, and the bearing vector $\mathbf{f}_{j}'$ of event. We parameterize the line with the rotation matrix $\mathbf{R}_{\ell}=[\mathbf{e}^{\ell}_{1} \, \mathbf{e}^{\ell}_{2} \, \mathbf{e}^{\ell}_{3}]$. Since scale is unobservable, we select the point $-\mathbf{e}^{\ell}_{3}$ at unit depth to lie on the line, and $\mathbf{e}^{\ell}_{1}$ to indicate its direction. Due to the aperture problem, we can only observe the projected camera velocity $\hat{\mathbf{v}}$ with components $u^{\ell}_{y}$ and $u^{\ell}_{z}$ in the $\mathbf{e}^{\ell}_{2}$ and $\mathbf{e}^{\ell}_{3}$ direction respectively.}
  \label{fig:geometry}
\end{figure}

%% file: figures/fig_solution_multiplicity.tex
\begin{figure*}
  \centering
  \begin{tabular}{cccc}
    \includegraphics[width=0.225\linewidth]{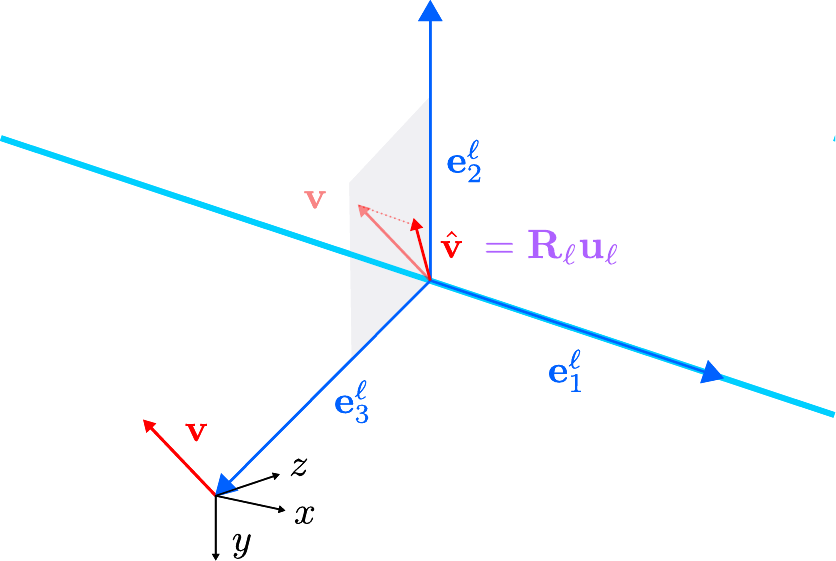} &
    \includegraphics[width=0.225\linewidth]{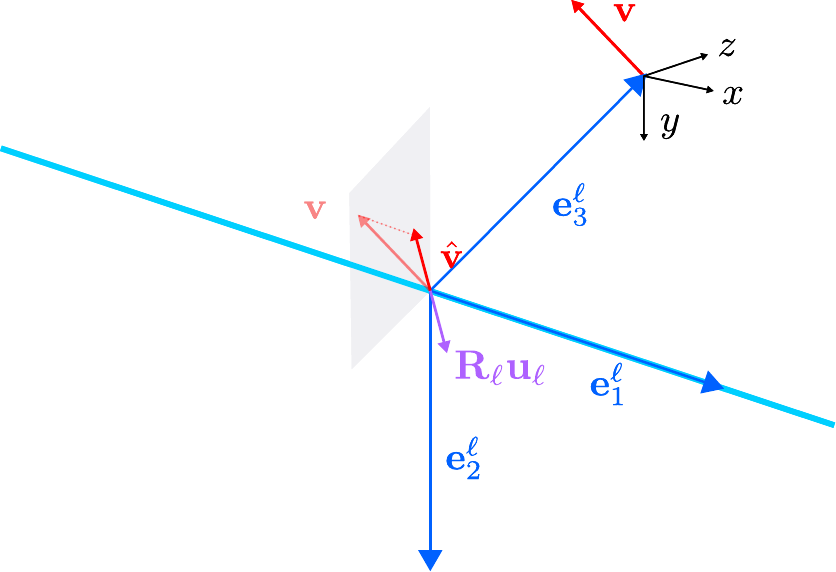} &
    \includegraphics[width=0.225\linewidth]{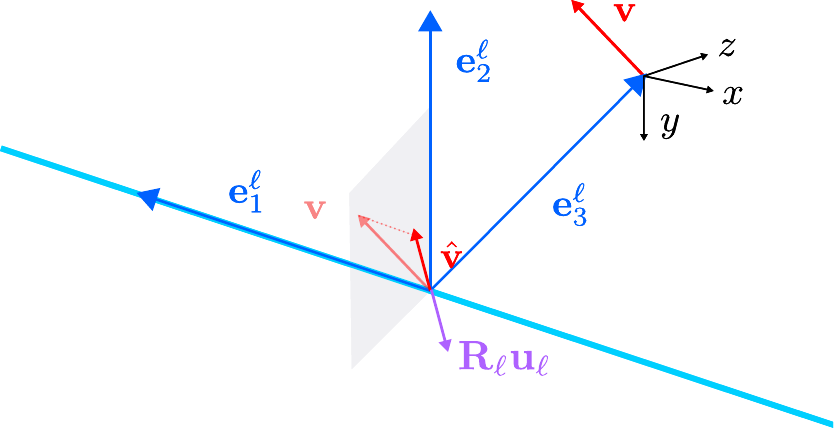} &
    \includegraphics[width=0.225\linewidth]{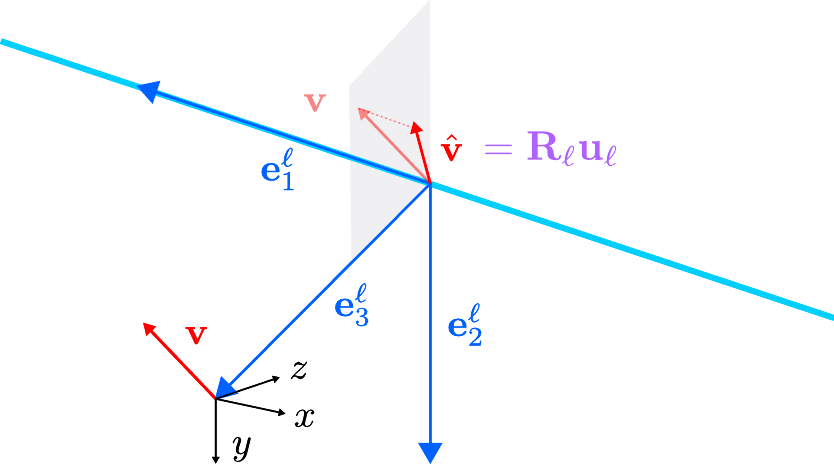} \\
    (a) Solution $S_0$ & (b) Solution $S_1$ & (c) Solution $S_2$ & (d) Solution $S_3$
  \end{tabular}
  \vspace{-1ex}
  \caption{Multiplicity of solutions to the incidence relation in~\eqref{eq:incidence_final}. While $S_0$ and $S_3$ have the line in front of the camera, $S_1$ and $S_2$ have the line behind the camera. The solution pairs $S_0, S_3$ and $S_1, S_2$ differ in the orientation of $\mathbf{e}^{\ell}_{1}$, which comes from the ambiguity of defining the line direction. In solutions with the line behind the camera, the measured projected camera velocity, $\mathbf{R}_\ell \mathbf{u}_\ell$ returned by the solver is the negative of the true projected velocity $\hat{\mathbf{v}}$. However, these solutions can be discarded by enforcing the condition in~\eqref{eq:solution_checking}.}
  \label{fig:solution_multiplicity}
  \vspace{-2ex}
\end{figure*}

%% file: figures/fig_velocity_averaging.tex
\begin{figure}
  \centering
  \resizebox{\linewidth}{!}{
  \begin{tabular}{cc}
    \includegraphics[height=3cm]{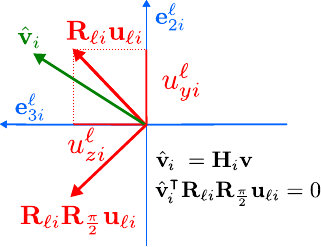} &
    \includegraphics[height=3cm]{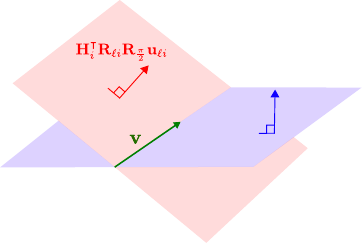}      \\
    (i) single line constraint & (ii) multiple line constraints
  \end{tabular}}
  \caption{(i) The line constraint in~\eqref{eq:line_constraint} dictates that the 90-degree rotated measured velocity $\mathbf{R}_{{\ell}i} \mathbf{R}_{\frac{\pi}{2}}\mathbf{u}_{{\ell}i}$ should be perpendicular to $\hat{\mathbf{v}}_{i}$, or $\mathbf{v}^\intercal \mathbf{H}_i^\intercal \mathbf{R}_{{\ell}i} \mathbf{R}_{\frac{\pi}{2}}\mathbf{u}_{{\ell}i} = 0$. (ii) Each such constraint spans a two-dimensional subspace, in which $\mathbf{v}$ must reside. With a minimum of two such subspaces, the velocity can be found.} 
  \label{fig:velocity_averaging}
\end{figure}

%% file: sec/4_implementation.tex
\section{Implementation}
\label{sec:implementation}

We integrate the aforementioned linear solver into a RANSAC framework for parameter determination of each manifold, followed by fitting over all inliers. Diverging from the approach of~\cite{gao2023eventail}, our implementation adopts the GC-RANSAC framework~\cite{barath2021graph} for robust geometric model estimation. GC-RANSAC enhances the original RANSAC by introducing a few versatile functionalities tailored for early termination, thereby expediting the overall process. Essentially, GC-RANSAC first iteratively selects a minimal, spatially coherent subset of events ($N = 5$) from the incoming event stream, applies the proposed linear solver described in~\algref{alg:linear_solver}, and evaluates the quality of the resulting hypothesis. Each occurrence of a \emph{so-far-the-best} hypothesis will trigger a local refinement within a subset of its inliers. This procedure is repeated $M$ times to separately identify $M$ manifolds. We will now delve into the critical aspects influencing this process.

\textbf{Spatially Coherent Sampler:} The manifold's continuous structure allows for the examination of the data's spatial coherency. We utilize NAPSAC~\cite{torr2002napsac} to sample from the incoming event stream. This approach starts by randomly selecting one event in the space-time volume followed by identifying four additional events within a hyper-sphere centered on the initial event, based on a predetermined radius $r$. In practice, these four points, located within the hyper-sphere, are likely to be inliers\footnote{Note that this method does not conflict with the findings about ensuring spatial distribution among samples~\cite{gao2023eventail}. However, we adhere to a general rule of maintaining this spatial distribution within our defined hyper-sphere.}. 

\textbf{Angular Reprojection Residual:} We employ the angular reprojection residual~\cite{lee2019closed} for inliers selection. The objective is to minimally correct the two lines, the bearing vector emanating from an event and the 3D line, so they could converge at a single point (\ie $\mathbf{P}_j$ in~\figref{fig:geometry}). Unlike the typical image reprojection residual, this measurement is invariant to both rotation and scale.

\textbf{Local Refinement:} RANSAC often results in numerous ineffective iterations. Therefore, when a promising hypothesis emerges, it is recommended to perform local refinement using the inlier sets, which can enhance the inlier ratio and decrease the total number of required iterations. In our work, we introduce two distinct methods for local refinement. The first method leverages the over-determined nature of our proposed linear solver, while the second employs non-linear optimization with Levenberg–Marquardt over an algebraic error (\ie~\eqref{eq:incidence_final}). As suggested in~\cite{barath2021graph}, we randomly select a subset from the inlier set ($N = 10$) and repeat this procedure $Q$ times. 

%% file: sec/5_experiments.tex
\section{Experiments}
\label{sec:experiments}

We perform evaluations both on synthetic and real data. We first confirm the runtime improvement and numerical stability of our linear solver. Next, we discuss the impact of the number of used events or lines, over different noise setups. We conclude with experiments on a few public real-world sequences, demonstrating the advantage over existing methods. We quantify the accuracy of our results with the same criterion as~\cite{gao2023eventail}, the direction error $\phi$ between the estimated and the ground truth velocities, given that the scale is not observable. All experiments are conducted on a 32GB RAM desktop with an Intel Core i9-10900F Processor.


\subsection{Simulation}
\label{sec:simulation}

We first evaluate the performance of the proposed linear solver under different setups over synthetic data. Readings from individual manifolds are generated as follows. We first sample randomly directed linear and angular velocities of \SI{0.5}{\meter/\second} and \SI{15}{\degree/\second} magnitude, respectively. The time window length is set to \SI{0.5}{\second}, and the virtual event camera has a resolution of 640$\times$480 and a focal length of 320 pixels. Next, we sample a random line in 3D with a finite length and sample random events according to the spatiotemporal strategy in~\cite{gao2023eventail}. We study three types of noise with different magnitudes: pixel noise (\SI{0.5}{\pixel}), timestamp jitter (\SI{0.5}{\ms}), and gyroscope noise (\SI{5.0}{\degree/\second}), which is assumed to be given by an IMU. The magnitude of the pixel noise and the noise on camera angular velocities are consistent within the same noise level but vary in direction. Timestamp noise follows a zero-mean Gaussian. For a more detailed sensitivity study for different noise sources and levels see the supplementary material. We generate one million random line, velocity, and event configurations, and report the mean and median angle error, as well as the minimum and mean runtime in milliseconds of our method and the one in~\cite{gao2023eventail}.

\input{tables/tab_sim_general_comparsion}

\textbf{Runtime Analysis:} In each run we record the runtime of both solvers. As reflected in~\tabref{tab:simulation_general_comparsion}, our solver runs over 600 times faster than the \Groebner basis solver in~\cite{gao2023eventail}. 

\textbf{Numerical Stability:} We further analyze the numerical stability of both methods under the noise-free setup. We report the likelihood of the solver to converge to within a low (\ang{1.0}, or \ang{0.1}) error. Our solver consistently reaches a zero error, unlike the polynomial solver, which, due to numerical instabilities of the elimination template, fails to converge within a \ang{0.1} error range \SI{1}{\percent} of times.

\input{tables/tab_sim_consistency_analysis.tex}
\input{tables/tab_real_world_exp.tex}
\input{figures/fig_sim_analysis_summary}

\textbf{Analysis of the Number of Used Events:} In each simulation, we sample 1,000 signal events and introduce a type of representative noise to the measurements. From these, we use the first $5 \leq K \leq 1,000$ events as input for our linear solver for a fair comparison and document the resulting error. We repeat this simulation a million times, varying $K$, and report the results in~\figref{fig:sim_analysis_summary}. We observe a clear trend that as the number of used events increases, the error decreases markedly, except when noise is introduced to the camera's angular velocity. This exception occurs because the solver cannot average out the noise on the angular velocity, regardless of the number of events processed. The other two errors approach near zero when 1,000 events are used. A full analysis of the solver's performance under each noise type can be found in the supplementary material.

\textbf{Analysis of the Number of Used Lines:} Finally, we validate our velocity averaging scheme. We extend our simulation to multiple manifolds. For each run, we sampled ten lines and, within each line, we selected ten signal events with known line associations as input to our solver. The first $2 \leq K \leq 10$ solutions (line and partial motion parameters) from each manifold were taken into the linear averaging scheme, and the error was recorded. This simulation was executed 10,000 times. \figref{fig:sim_analysis_summary} shows that as the number of used lines increases, the error drops significantly. A comprehensive analysis of the solver's performance against various noise types is available in the supplementary material. Additionally, we show quantitative results in~\tabref{tab:simulation_consistency_comparsion}. Both the \Groebner Solver and our linear solver use five lines with either five or ten events each. Our approach demonstrates a lower error with noisy measurements, and this margin grows further in overdetermined systems (\ie~$N = 10$).


\subsection{Real-world Experiment}
\label{sec:real_world_experiment}

Similarly to~\cite{gao2023eventail}, we validate our method on the same data sequence from VECtor Benchmark~\cite{gao2022vector}. Unlike the previous work, we first segmented the event data into non-overlapping intervals of \SI{0.1}{\s} each and reduced the overall size of the data to approximately 5,000 events per interval for efficiency. Motion-corrected bearing vectors are then calculated by fusing gyroscope readings from the attached IMU. Next, to construct the chosen sampler used in GC-RANSAC, we multiply the timestamp by a scale factor of 1,000 and established a radius of 50 to compose the spatially coherent graph in the space-time volume. We apply an angular reprojection threshold of \ang{0.2} for inlier selection, consistent across both the primary iterations and the local refinement stages. The number of iterations for each manifold fitting is capped at 100 and evaluated manifolds is capped at 10. In~\tabref{tab:real_world_experiment}, we summarized the performance, including both mean and median errors, across two baseline approaches and three variants of our proposed method. Importantly, as~\cite{gao2023eventail} reports, CELC+opt~\cite{peng2021continuous} is limited to certain sub-sequences, where spatial-temporal plane clustering is feasible, while, \cite{gao2023eventail} does not suffer from this limitation. We test three configurations of our method: \emph{linear only}, \emph{linear with non-minimal refinement}, and \emph{linear with non-linear optimization}. Each configuration uses GC-RANSAC with the spatially coherent sampler. The latter two configurations perform different operations when a new best hypothesis is found by the sampler: ``Linear w/ non-min. solver" samples 10 events from the found inliers and feeds them to the linear solver, resulting in a refined solution. ``Linear w/ non-linear opt." runs on-manifold Levenberg-Marquardt optimization steps (with line parameters $\mathbf{R}_\ell\in SO(3)$) over the 10 selected events and minimizes the algebraic error in~\eqref{eq:incidence_final}. We use a subset of 10 inliers to enhance efficiency and reduce errors, as seen in simulated experiments.

\textbf{Results:} In each configuration, our method achieves a lower error than the two baseline approaches. Additionally, local refinement enhances accuracy significantly. Without refinement it already has a \SI{10}{\percent} lower mean error than~\cite{peng2021continuous}, and \SI{24}{\percent} lower mean error than~\cite{gao2023eventail}. Introducing non-minimal solver refinement reduces the mean error by another \SI{18}{\percent} over the ``Linear only" baseline and non-linear optimization reduces it by \SI{16}{\percent}.

%% file: tables/tab_sim_general_comparsion.tex
\begin{table}[!t]
	\caption{General comparison between the two solvers.}
	\label{tab:simulation_general_comparsion}
	\footnotesize
	\begin{center}
		\vspace{-1.5ex}
		\begin{tabular}{l|cc|cc}
			\toprule  
			\multicolumn{1}{c|}{\multirow{2}{*}{Method}} & \multicolumn{2}{c|}{Runtime (\SI{}{\us})} & \multicolumn{2}{c}{Error Rate (\%)} \\
			\multicolumn{1}{c|}{}                        & min.              & avg.                  & $\geq \ang{0.1}$ & $\geq \ang{1.0}$ \\ 
			\midrule 
			\Groebner~\cite{gao2023eventail}             & 1893              & 2046                  & 1.00             & 0.28             \\
			\textbf{Linear (ours)}                       & \textbf{3.00}     & \textbf{3.25}         & \textbf{0.00}    & \textbf{0.00}    \\ 
			\bottomrule
		\end{tabular}
		\vspace{-2ex}
	\end{center}
\end{table}

%% file: tables/tab_sim_consistency_analysis.tex
\begin{table}[!t]
  \caption{Noise resilience of our linear solver and the solver in~\cite{gao2023eventail}. Left corresponds to the mean, right to the median error in degrees.}
  \label{tab:simulation_consistency_comparsion}
  \footnotesize
  \begin{center}
    \vspace{-1.5ex}
    \resizebox{0.98\linewidth}{!}{%
    \begin{tabular}{c|c|c|c|c}
      \toprule
      \multirow{2}{*}{Method}             & num.   & Pixel Noise        & Time. Jitter       & Gyro. Noise                 \\     
                                          & events & (\SI{0.5}{\pixel}) & (\SI{0.5}{\ms})    & (\SI{5.0}{\degree/\second}) \\
      \midrule
      \Groebner~\cite{gao2023eventail}    & 5      & 7.80/1.67          & 3.61/0.83          & 7.48/3.09                   \\
      \textbf{Linear (ours)}              & 5      & 5.53/1.24          & 2.87/0.73          & 6.53/2.47                   \\
      \textbf{Linear (ours)}              & 10     & \textbf{0.46/0.15} & \textbf{0.17/0.12} & \textbf{1.50/1.17}          \\
      \bottomrule
    \end{tabular}}
    \vspace{-2ex}
  \end{center}
\end{table}

%% file: tables/tab_real_world_exp.tex
\begin{table*}[!t]
  \caption{Real-world experiment results. Left corresponds to the mean, right to the median error in degrees.}
  \label{tab:real_world_experiment}
  \footnotesize
  \begin{center}
    \vspace{-1.5ex}
    \resizebox{0.95\linewidth}{!}{%
    \begin{tabular}{c|c|c|c|c|c}
      \toprule
      Seq. Name                  & CELC+opt~\cite{peng2021continuous} & \Groebner~\cite{gao2023eventail} & \textbf{Linear only}    & \textbf{Linear w/ non-min. solver} & \textbf{Linear w/ non-linear opt.} \\
      \midrule
      \emph{mountain-normal} & 27.7/29.3 & 33.5/33.6 & 25.2/21.4 & 17.0/17.2 & \textbf{16.5/14.6} \\
      \emph{desk-normal}     & 26.6/26.6 & 26.4/26.7 & 22.7/23.4 & \textbf{19.8/19.2} & 22.1/20.7 \\
      \emph{sofa-normal}     & 24.0/26.1 & 31.5/29.5 & 21.9/17.6 & 20.6/16.1 & \textbf{19.9/15.0} \\
      \bottomrule
    \end{tabular}}
    \vspace{-2ex}
  \end{center}
\end{table*}

%% file: figures/fig_sim_analysis_summary.tex
\begin{figure*}[t]
  \centering
  \begin{subfigure}{0.48\linewidth}
    \centering
    \includegraphics[width=\linewidth]{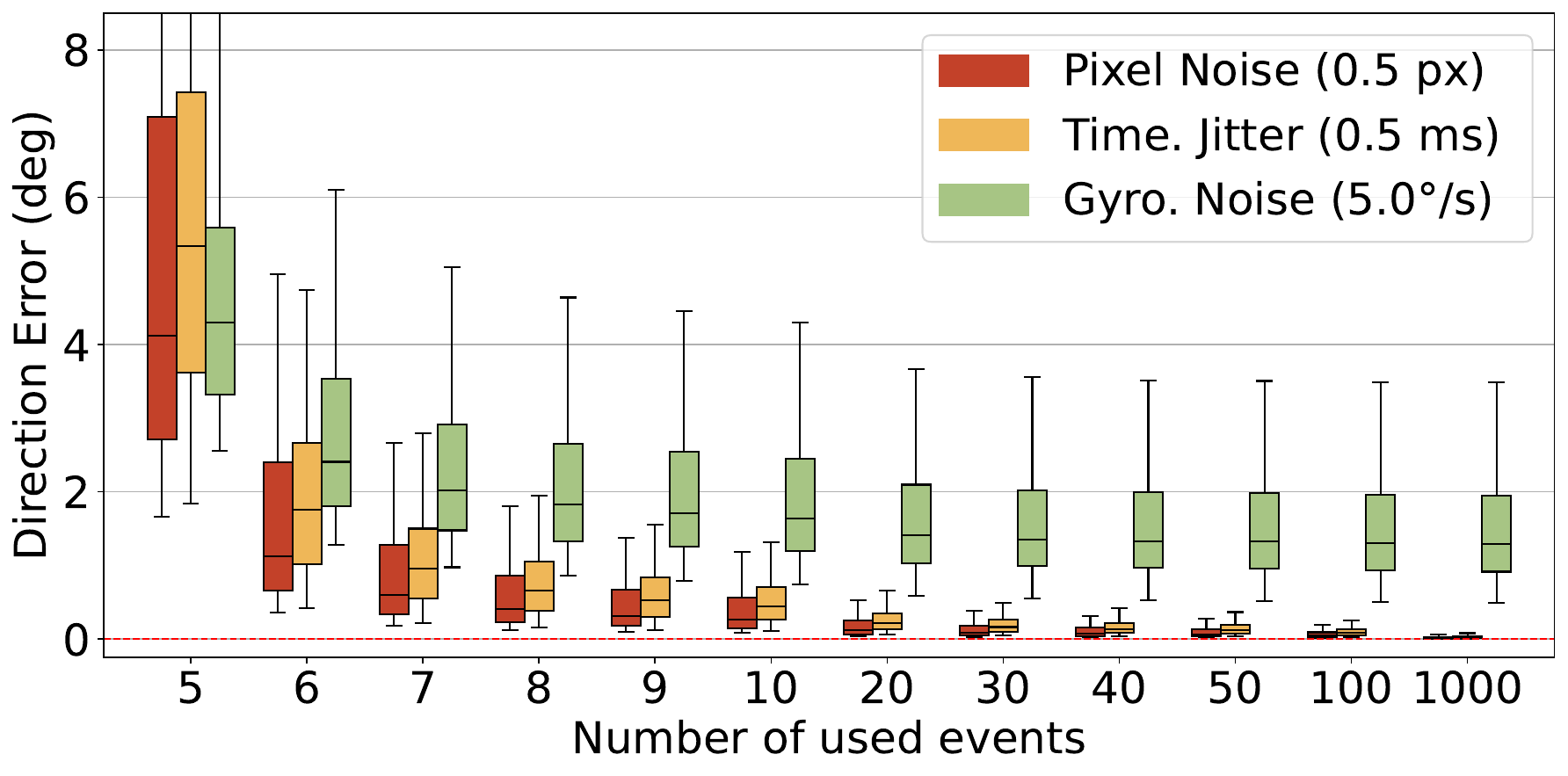}
  \end{subfigure}
  \hfill
  \begin{subfigure}{0.48\linewidth}
    \centering
    \includegraphics[width=\linewidth]{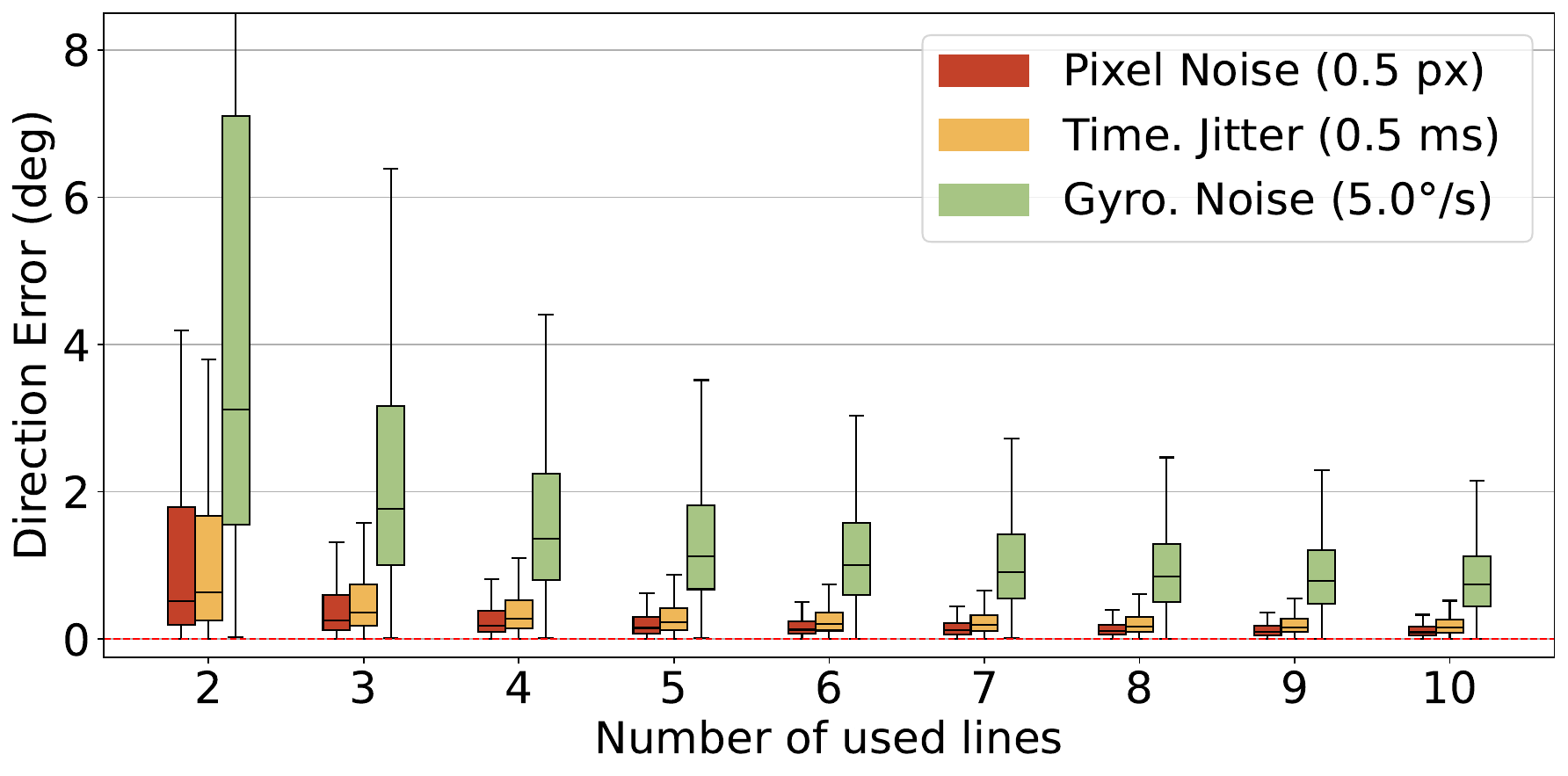}
  \end{subfigure}
  \caption{Condensed analysis of the number of used events (left) and used lines (right) over three types of representative noise.}
  \label{fig:sim_analysis_summary}
  \vspace{-1ex}
\end{figure*}

%% file: sec/6_conclusion.tex
\section{Conclusion and Future Work}
\label{sec:conclusion}

This work introduced a novel, efficient, and linear N-point solver for line-based relative motion estimation of an event camera. Compared to existing works that rely on polynomial system solvers, our method is more numerically stable, over 600 times faster, and allows the identification of degenerate cases explicitly. Moreover, we introduce a novel velocity averaging scheme that is simpler and faster than previous approaches. When combined with GC-RANSAC we show improved normalized velocity estimation compared to existing approaches, at a fraction of the runtime. Finally, the solutions found by our solver deliver new insights into event manifolds generated by lines and thus pave the way for line-based motion estimation with events. Moreover, despite focusing on event cameras in this work, our formulation is fully compatible with line detections from standard cameras. Thus the tools developed in this work can benefit both frame- and event-based computer vision. Our next steps consist of adding uncertainties to the partial velocity readings and applying the fusion strategy asynchronously over time as well as in conjunction with IMU measurements.

%% file: sec/7_acknowledgments.tex
\section*{Acknowledgments}
\label{sec:acknowledgments}

This research has been supported by projects 22DZ1201900 and 22ZR1441300 funded by the Natural Science Foundation of Shanghai as well as project 62250610225 by the National Science Foundation of China (NSFC). This work was also supported by the European Research Council (ERC) under grant agreement No. 864042 (AGILEFLIGHT).

%% file: sec/X_suppl.tex
\clearpage
\maketitlesupplementary

\section{Appendix}

Here we report additional results of our algorithm for varying noise sources in~\secref{sec:app:noise}, before discussing the proofs of Theorem 1 and Theorem 2 in~\secref{sec:app:thm1} and \secref{sec:app:thm2}, as well as the connection of our proposed line averaging scheme with that of~\cite{gao2023eventail} in~\secref{sec:app:linavg}. Finally, we provide additional visual insights into the manifolds spanned by events generated by a line. We show that these manifolds can be canonicalized, \ie~reduced to a small family of manifolds which are highly interpretable (see~\secref{sec:app:manifold}). 


\subsection{Noise Sensitivity Analysis}
\label{sec:app:noise}

In~\figref{fig:sim_event_number_analysis}, we provide additional results of our method in simulation, as we vary the number of events used by our solver, and the magnitude of the various noise sources, \eg~pixel noise, timestamp jitter, gyroscope noise. As expected, we see that all errors decrease as more events are used, and errors increase as more noise is injected. Again, the only noise source that cannot be completely eliminated through addition of events is the gyroscope noise, which introduces systematic errors. Experimentally, we found that $N=10$ events gives a good tradeoff between the speed of the algorithm, and observed errors for all noise levels and sources. 

We also present additional results for differing noise sources and magnitudes of our line averaging scheme in~\figref{fig:sim_multiple_clusters_analysis}, and analyse the resulting errors as the number of used lines increases. Again we see that all errors tend to zero as more lines are used, except for the gyroscope noise.


\subsection{Proof of Theorem 1 on Degeneracies}
\label{sec:app:thm1}

For clarity, we restate Theorem 1 here: \\

Theorem 1: \noindent\emph{If $\text{rank}(\mathbf{A})\geq 5$, with $\mathbf{A}$ defined in~\eqref{eq:linear_system}, the decomposition in Eqs.~(\ref{eq:def_scale},~\ref{eq:solver_1},~\ref{eq:decomp_e1}) always succeeds and yields four distinct solutions. If $\text{rank}(\mathbf{A})< 5$ the solver returns infinitely many solutions.} \\

\textbf{Proof: }
First assume $\text{rank}(\mathbf{A})\geq 5$. Then SVD returns two distinct principle directions $\pm \hat{\mathbf{x}}$. After decomposition, \eqref{eq:decomp_e1} yields two more solutions, resulting in a total of four distinct solutions. Now assume that the decomposition fails, and this can happen for three reasons: 

\emph{Failure to normalize in~\eqref{eq:def_scale}:} Normalization may fail if $\hat{\mathbf{x}}_{4:6}$ has zero norm. However, this case is impossible for a matrix $\mathbf{A}$ with rank $\geq 5$ for the following reason: Let $\mathbf{B},\mathbf{C}$ be the three left and right columns of $\mathbf{A}$ (see~\eqref{eq:linear_system}). Moreover, note that $\mathbf{C}=\mathbf{T}\mathbf{B}$, where $\mathbf{T}=\text{diag}(t'_1, t'_2, ..., t'_N)$ is a diagonal matrix, \ie~each row of $\mathbf{B}$ is a multiple of the corresponding row in $\mathbf{C}$.

If $\hat{\mathbf{x}}_{4:6}$ has zero norm, $\hat{\mathbf{x}}_{4:6} = 0$. Next, let $\sigma$ be the smallest singular value of $\mathbf{A}$ corresponding to the solution $\hat{\mathbf{x}}$. Then 
\begin{align}
  \mathbf{A}^\intercal\mathbf{A}\hat{\mathbf{x}}&=\sigma \hat{\mathbf{x}}\\
  \begin{bmatrix}
    \mathbf{B}^\intercal\\\mathbf{B}^\intercal \mathbf{T} 
  \end{bmatrix}
  \begin{bmatrix}
    \mathbf{B}&\mathbf{T}\mathbf{B}
  \end{bmatrix}
  \begin{bmatrix}
    \hat{\mathbf{x}}_{1:3}\\0
  \end{bmatrix}&=\begin{bmatrix}
    \sigma\hat{\mathbf{x}}_{1:3}\\0
  \end{bmatrix}\\
  \begin{bmatrix}
    \mathbf{B}^\intercal\mathbf{B}\hat{\mathbf{x}}_{1:3}\\
    \mathbf{B}^\intercal \mathbf{T}\mathbf{B}\hat{\mathbf{x}}_{1:3}
  \end{bmatrix}&=\begin{bmatrix}
    \sigma\hat{\mathbf{x}}_{1:3}\\0
  \end{bmatrix}
\end{align}
The last three rows of the last equation are 
\begin{align}
  \mathbf{B}^\intercal \mathbf{T}\mathbf{B}\hat{\mathbf{x}}_{1:3} = 0 \,,
\end{align}
and imply either that $\mathbf{B}^\intercal(\mathbf{T}\mathbf{B}\hat{\mathbf{x}}_{1:3})=0$, \ie~$\mathbf{T}\mathbf{B}\hat{\mathbf{x}}_{1:3}$ is in the left null-space of $\mathbf{B}$, or   $\mathbf{B}\hat{\mathbf{x}}_{1:3}=0$, \ie~$\hat{\mathbf{x}}_{1:3}$ is in the right null-space of $\mathbf{B}$. Both imply that $\text{rank}(\mathbf{B}) < 3$. This can only be the case if $\text{rank}(\mathbf{A}) = 5$, following the assumption. This implies that the smallest singular value is $\sigma=0$. From the first three equations above, this implies that $\mathbf{B}\hat{\mathbf{x}}_{1:3}=0$. But then 
\begin{align}
  \mathbf{A}\begin{bmatrix}
    0\\\mathbf{x}_{1:3}
  \end{bmatrix}=\mathbf{T}\mathbf{B}\mathbf{x}_{1:3}=0 \,,
\end{align}
which implies that $\hat{\mathbf{x}}=[0^\intercal\,\mathbf{x}_{1:3}^\intercal]$ is also in the null space of $\mathbf{A}$. We now find that both $\hat{\mathbf{x}}_1=[0^\intercal\,\mathbf{x}_{1:3}^\intercal]$ and $\hat{\mathbf{x}}_2=[\mathbf{x}_{1:3}^\intercal\,0^\intercal]$ are in the null-space of $\mathbf{A}$. These vectors are independent, and render the rank of $\mathbf{A}<5$. This is a contradiction.

\emph{Failure to recover $\mathbf{e}^{\ell}_{1}$:} Recovering $\mathbf{e}^{\ell}_{1}$ fails if the norm of the cross product in~\eqref{eq:decomp_e1} is $0$. This implies that $\hat{\mathbf{x}}_{4:6} = \lambda \hat{\mathbf{x}}_{1:3}$. For similar reasons as above, this implies that $\hat{\mathbf{x}}_{1:3}$ solves both $\mathbf{B}\mathbf{x} = 0$ and $\mathbf{C}\mathbf{x} = 0$. This implies that $\lambda$ can be freely varied, which would imply a two-dimensional null space of $\mathbf{A}$ and a rank $\leq 4$ which is again a contradiction.

\emph{Line passing through the origin at $t'=0$:} Note that in such a case, $\mathbf{e}_3^\ell$ would not be defined, and could cause issues in solving. However, we can then use a different definition of the line, with the direction $\mathbf{d}=\mathbf{e}_1^\ell$, and point on the line $\mathbf{P}=\mathbf{e}_1^\ell$. The line moment then becomes $\mathbf{m}=\mathbf{P}\times \mathbf{d}=0$. Inserting this into~\eqref{eq:nonminimal_incidence}, transforms~\eqref{eq:incidence_final} into
\begin{equation}
  {\mathbf{f}_j'}^\intercal (\mathbf{e}_3^\ell u_y^\ell - \mathbf{e}_2^\ell u_z^\ell)=0 \,.
\end{equation}
However, this would imply that the system in \eqref{eq:linear_system} has a solution of the form $\hat{\mathbf{x}}=[\hat{\mathbf{x}}_{1:3}^\intercal\,\mathbf{0}^\intercal]^\intercal$, with $\hat{\mathbf{x}}_{1:3}=\mathbf{e}_3^\ell u_y^\ell - \mathbf{e}_2^\ell u_z^\ell$. However, we proved in the last two cases that such a solution form implies that the rank of $\mathbf{A}$ is smaller than 5. Thus ensuring $\text{rank}(\mathbf{A})\geq 5$ is sufficient for discarding the case where the line passes through the origin.

\input{figures/fig_sim_event_number_analysis}

\input{figures/fig_sim_multiple_clusters_analysis}

We conclude that if $\text{rank}(\mathbf{A})\geq 5$, the decomposition cannot fail, and always returns four distinct solutions. Moreover, we conclude that a $\text{rank}(\mathbf{A}) < 5$ yields solutions $\hat{\mathbf{x}}$ from a two dimensional nullspace, which yields infinitely many decompositions. \hfill $\blacksquare$


\subsection{Proof of Theorem 2 on Solution Multiplicity}
\label{sec:app:thm2}

For clarity, we restate Theorem 2 here: \\

Theorem 2: \emph{Given a solution $S_0=\{\mathbf{e}^{\ell}_{1}, \mathbf{e}^{\ell}_{2}, \mathbf{e}^{\ell}_{3}, u^{\ell}_{y}, u^{\ell}_{z}\}$ to the incidence relation in~\eqref{eq:incidence_final}, then 
\begin{align}
  \nonumber S_1 &= \{\mathbf{e}^{\ell}_{1}, -\mathbf{e}^{\ell}_{2}, -\mathbf{e}^{\ell}_{3}, u^{\ell}_{y}, u^{\ell}_{z}\} \,, \\
  \nonumber S_2 &= \{-\mathbf{e}^{\ell}_{1}, \mathbf{e}^{\ell}_{2}, -\mathbf{e}^{\ell}_{3}, -u^{\ell}_{y}, u^{\ell}_{z}\} \,, \\
  \nonumber S_3 &= \{-\mathbf{e}^{\ell}_{1}, -\mathbf{e}^{\ell}_{2}, \mathbf{e}^{\ell}_{3}, -u^{\ell}_{y}, u^{\ell}_{z}\}
\end{align}
are also solutions. These four solutions are visualized in ~\figref{fig:solution_multiplicity}. For solutions $S_1$ and $S_2$ the closest point $-\mathbf{e}_3^\ell$ on the line is behind the camera, while for solutions $S_2$ and $S_3$ the line direction $\mathbf{e}^{\ell}_{1}$ is flipped, which represents an ambiguity in the definition of direction of $\mathbf{L}$.} \\

\textbf{Proof:} 
We will only prove solutions $S_1$ and $S_2$ since $S_3$ can be derived from a composition of $S_1$ and $S_2$. Inserting $S_1 = \{{\mathbf{e}^{\ell}_{1}}', {\mathbf{e}^{\ell}_{2}}', {\mathbf{e}^{\ell}_{3}}', {u^{\ell}_{y}}', {u^{\ell}_{z}}'\} = \{\mathbf{e}^{\ell}_{1}, -\mathbf{e}^{\ell}_{2}, -\mathbf{e}^{\ell}_{3}, u^{\ell}_{y}, u^{\ell}_{z}\}$ into~\eqref{eq:incidence_final}, we have
\begin{align}
  \nonumber  & \ \ \ t_{j}'{\mathbf{f}_{j}'}^\intercal ({u^{\ell}_{z}}' {\mathbf{e}^{\ell}_{2}}' - {u^{\ell}_{y}}' {\mathbf{e}^{\ell}_{3}}') + {\mathbf{f}_{j}'}^\intercal {\mathbf{e}^{\ell}_{2}}'\\
  \nonumber =& \ \ \ t_{j}'{\mathbf{f}_{j}'}^\intercal (u^{\ell}_{z} (-\mathbf{e}^{\ell}_{2}) - u^{\ell}_{y} (-\mathbf{e}^{\ell}_{3})) + {\mathbf{f}_{j}'}^\intercal (-\mathbf{e}^{\ell}_{2}) \\
  \nonumber =& \ \ \ -(t_{j}'{\mathbf{f}_{j}'}^\intercal (u^{\ell}_{z} \mathbf{e}^{\ell}_{2} - u^{\ell}_{y} \mathbf{e}^{\ell}_{3}) + {\mathbf{f}_{j}'}^\intercal \mathbf{e}^{\ell}_{2}) \\
  \nonumber =& \ \ \ 0 \hspace{7.4cm}
\end{align}
and $S_2 = \{{\mathbf{e}^{\ell}_{1}}'', {\mathbf{e}^{\ell}_{2}}'', {\mathbf{e}^{\ell}_{3}}'', {u^{\ell}_{y}}'', {u^{\ell}_{z}}''\} = \{-\mathbf{e}^{\ell}_{1}, \mathbf{e}^{\ell}_{2}, -\mathbf{e}^{\ell}_{3}, -u^{\ell}_{y}, u^{\ell}_{z}\}$ yields 
\begin{align}
  \nonumber  & \ \ \ t_{j}'{\mathbf{f}_{j}'}^\intercal ({u^{\ell}_{z}}'' {\mathbf{e}^{\ell}_{2}}'' - {u^{\ell}_{y}}''{\mathbf{e}^{\ell}_{3}}'') + {\mathbf{f}_{j}'}^\intercal {\mathbf{e}^{\ell}_{2}}''\\
  \nonumber =& \ \ \ t_{j}'{\mathbf{f}_{j}'}^\intercal (u^{\ell}_{z} \mathbf{e}^{\ell}_{2}-(-u^{\ell}_{y})(-\mathbf{e}^{\ell}_{3})) + {\mathbf{f}_{j}'}^\intercal \mathbf{e}^{\ell}_{2}\\
  \nonumber =& \ \ \ t_{j}'{\mathbf{f}_{j}'}^\intercal (u^{\ell}_{z} \mathbf{e}^{\ell}_{2}-u^{\ell}_{y}\mathbf{e}^{\ell}_{3}) + {\mathbf{f}_{j}'}^\intercal \mathbf{e}^{\ell}_{2}\\
  \nonumber =& \ \ \ 0 \hspace{7.4cm} 
\end{align}
\hfill $\blacksquare$


\subsection{Handling of Parallel Lines}

As mentioned in the main text, parallel lines may cause difficulties in identifying the $\mathbf{e}_1^\ell$ direction of the camera velocity. However, we can identify this case easily by checking the rank of $\mathbf{D}$. If it is lower than 2, we can discard the sample, and select a new one, or even use another RANSAC loop to select pairs of lines until the rank of $\mathbf{D}$ is at least 2. Let us now prove that parallel lines cause a rank deficiency in $\mathbf{D}$.

\textbf{Proof:}
We will proceed in showing that if two lines are parallel, the two corresponding rows $\mathbf{r}_1=u_{y1}^\ell \mathbf{e}_{31}^\ell-u_{z1}^\ell \mathbf{e}_{21}^\ell$ and $\mathbf{r}_2=u_{y2}^\ell \mathbf{e}_{32}^\ell-u_{z2}^\ell \mathbf{e}_{22}^\ell$ in $\mathbf{D}$ are parallel and will thus result in rank deficiency (see~\eqref{eq:velocity_averaging}). Expanding $\mathbf{v}$ in the two line coordinate frames yields
\begin{align}
  \lambda_1\mathbf{v}=u_{x1}^\ell\mathbf{e}_{11}^\ell+u_{y1}^\ell\mathbf{e}_{21}^\ell+u_{z1}^\ell \mathbf{e}_{31}^\ell\\
  \lambda_2\mathbf{v}=u_{x2}^\ell\mathbf{e}_{12}^\ell+u_{y2}^\ell\mathbf{e}_{22}^\ell+u_{z2}^\ell \mathbf{e}_{32}^\ell
\end{align}
with unknown scale factors $\lambda_1,\lambda_2$. For parallel lines $\mathbf{e}^\ell_{11}=\mathbf{e}^\ell_{12}\doteq\mathbf{e}^\ell_{1}$. Computing $\mathbf{e}^\ell_{1}\times \mathbf{v}$ in two ways (with two expansions of $\mathbf{v})$, we recover exactly the rows of $\mathbf{D}$ by
\begin{align}
  \lambda_1 (\mathbf{e}^\ell_{1}\times\mathbf{v})=u_{y1}^\ell \mathbf{e}_{31}^\ell-u_{z1}^\ell \mathbf{e}_{21}^\ell=\mathbf{r}_1 \\
  \lambda_2 (\mathbf{e}^\ell_{1}\times\mathbf{v})=u_{y2}^\ell \mathbf{e}_{32}^\ell-u_{z2}^\ell \mathbf{e}_{22}^\ell=\mathbf{r}_2 
\end{align}
It follows that $\mathbf{r}_1=\frac{\lambda_1}{\lambda_2} \mathbf{r}_2$, \ie~they are parallel. \hfill $\blacksquare$


\subsection{Global Optimality of \texorpdfstring{$\mathbf{R}_l$}{TEXT} and \texorpdfstring{$\mathbf{u}_l$}{TEXT}}

As noted in the main text, while the SVD-based solver which recovers $\hat{\mathbf{x}}$ from a set of incidence relations (\eqref{eq:linear_system}) finds a globally optimal solution $\hat{\mathbf{x}}$, it is not clear if the decomposed solution $\mathbf{R}_l,\mathbf{u}_l$ is also optimal with respect to the same objective. We prove this here.\\ 

\textbf{Proof:} 
We will prove this by way of contradiction. Assume given the SVD-based solution 
\begin{align}
  \nonumber\hat{\mathbf{x}}=\arg\min_{\mathbf{x}} \Vert \mathbf{A}\mathbf{x}\Vert^2\quad
  \nonumber\text{such that}\quad \Vert\mathbf{x}\Vert^2=1.
\end{align}
which is globally optimal, and decomposition $\mathbf{R}_l,\mathbf{u}_l$ with $\hat{\mathbf{x}}=\mathbf{x}(\mathbf{R}_l,\mathbf{u}_l)$. Assume that there exists a different, more optimal $\mathbf{R}'_l,\mathbf{u}'_l$ with $\hat{\mathbf{x}}'=\mathbf{x}(\mathbf{R}'_l,\mathbf{u}'_l)$. Then $\Vert \mathbf{A}\hat{\mathbf{x}}'\Vert^2 < \Vert \mathbf{A}\hat{\mathbf{x}}\Vert^2$ but this is impossible since it would imply that $\hat{\mathbf{x}}'$ is more optimal than $\hat{\mathbf{x}}$, but $\hat{\mathbf{x}}$ is already optimal. This implies that the objective is already optimal in $\mathbf{R}_l,\mathbf{u}_l$ which concludes the proof. \hfill $\blacksquare$


\subsection{Connection between the Proposed Line Averaging Scheme and \cite{gao2023eventail}}
\label{sec:app:linavg}

The presented velocity averaging scheme is conceptually simpler, and lends itself to geometric interpretation, unlike the scheme in~\cite{gao2023eventail}. However, surprisingly these schemes are actually equivalent, as will be demonstrated next. In~\cite{gao2023eventail}, \eqref{eq:line_constraint} is used to set up a number of constraints 
\begin{align}
  {\mathbf{e}_{2i}^\ell}^\intercal \mathbf{v} - \lambda_i u_{yi}^\ell = 0\\
  {\mathbf{e}_{3i}^\ell}^\intercal \mathbf{v} - \lambda_i u_{zi}^\ell = 0
\end{align}
Introducing unknowns $\mathbf{v}$ and $\{\lambda_i\}$, one for each line. Stacking multiple of these equations results in a system 
\begin{equation}
  \underbrace{\begin{bmatrix}
    {\mathbf{e}_{21}^\ell}^\intercal&-u_{y1}^\ell & \cdots&0\\
    {\mathbf{e}_{31}^\ell}^\intercal&-u_{z1}^\ell&\cdots&0\\
    \vdots&\ddots&\ddots&\vdots\\
    {\mathbf{e}_{2M}^\ell}^\intercal&0&\cdots&-u_{yM}^\ell\\
    {\mathbf{e}_{3M}^\ell}^\intercal&0&\cdots&-u_{zM}^\ell\\
  \end{bmatrix}}_{\mathbf{E}}\begin{bmatrix}
    \mathbf{v}\\ \lambda_1\\\vdots\\\lambda_M
  \end{bmatrix}=\mathbf{0} \,.
\end{equation}
This system is then multiplied from the left with $\mathbf{E}^\intercal$, and the Shur complement trick is employed to eliminate the extraneous variables $\lambda_i$, resulting in the equation $\mathbf{F} \mathbf{v} = \mathbf{0}$, where we use the following definitions:
\begin{align}
  \mathbf{F} &= \mathbf{U} - \mathbf{W}\mathbf{V}^{-1}\mathbf{W}^\intercal\\
  \mathbf{U} &= \sum_{i=1}^M (\mathbf{e}^\ell_{2i}{\mathbf{e}^\ell_{2i}}^\intercal + \mathbf{e}^\ell_{3i}{\mathbf{e}^\ell_{3i}}^\intercal)\\
  \mathbf{V} &= \text{diag}\left({u_{y1}^\ell}^2+{u_{z1}^\ell}^2, \cdots, {u_{yM}^\ell}^2+{u_{zM}^\ell}^2\right)\\
  \mathbf{W}^\intercal &= \begin{bmatrix}
    -u_{y1}^\ell{\mathbf{e}_{21}^\ell}^\intercal-u_{z1}^\ell{\mathbf{e}_{31}^\ell}^\intercal\\
    \vdots\\
    -u_{yM}^\ell{\mathbf{e}_{2M}^\ell}^\intercal-u_{zM}^\ell{\mathbf{e}_{3M}^\ell}^\intercal\\
  \end{bmatrix}
\end{align}
Inserting the equations, and simplifying we get 
\begin{align}
  \mathbf{F} =& \sum_{i=1}^M (\mathbf{e}^\ell_{2i}{\mathbf{e}^\ell_{2i}}^\intercal + \mathbf{e}^\ell_{3i}{\mathbf{e}^\ell_{3i}}^\intercal) \nonumber \\ 
  &- \frac{\left(-u_{yi}^\ell\mathbf{e}_{2i}^\ell-u_{zi}^\ell\mathbf{e}_{3i}^\ell\right)\left(-u_{yi}^\ell{\mathbf{e}_{2i}^\ell}^\intercal-u_{zi}^\ell{\mathbf{e}_{3i}^\ell}^\intercal\right)}{{u_{yi}^\ell}^2+{u_{zi}^\ell}^2}\\
  =&\sum_{i=1}^M \frac{1}{{u_{yi}^\ell}^2+{u_{zi}^\ell}^2}\left({u_{zi}^\ell}^2\mathbf{e}_{2i}^\ell{\mathbf{e}_{2i}^\ell}^\intercal+{u_{yi}^\ell}^2\mathbf{e}_{3i}^\ell{\mathbf{e}_{3i}^\ell}^\intercal \right. \nonumber \\
  &- \left. u_{yi}^\ell u_{zi}^\ell\mathbf{e}^\ell_{3i}{\mathbf{e}^\ell_{2i}}^\intercal - u_{yi}^\ell u_{zi}^\ell\mathbf{e}^\ell_{2i}{\mathbf{e}^\ell_{3i}}^\intercal \right)\\
  =&\sum_{i=1}^M \frac{\left(u_{yi}^\ell \mathbf{e}_{3i}^\ell - u_{zi}^\ell \mathbf{e}_{2i}^\ell\right)\left(u_{yi}^\ell \mathbf{e}_{3i}^\ell - u_{zi}^\ell \mathbf{e}_{2i}^\ell\right)^\intercal}{{u_{zi}^\ell}^2+{u_{zi}^\ell}^2}\\
  =&\hat{\mathbf{D}}^\intercal\hat{\mathbf{D}}
\end{align}
with 
\begin{align}
  \hat{\mathbf{D}} = \begin{bmatrix}
    \frac{1}{\sqrt{{u_{y1}^\ell}^2+{u_{z1}^\ell}^2}}\left(u_{y1}^\ell \mathbf{e}_{31}^\ell - u_{z1}^\ell \mathbf{e}_{21}^\ell\right)\\
    \vdots\\
    \frac{1}{\sqrt{{u_{yM}^\ell}^2+{u_{zM}^\ell}^2}}\left(u_{yM}^\ell \mathbf{e}_{3M}^\ell - u_{zM}^\ell \mathbf{e}_{2M}^\ell\right)\\
  \end{bmatrix}
\end{align}
The linear system thus becomes 
\begin{align}
  \mathbf{F}\mathbf{v} &= \mathbf{0}\\
  \hat{\mathbf{D}}^\intercal \hat{\mathbf{D}}\mathbf{v} &= \mathbf{0}\\
  \hat{\mathbf{D}}^\intercal (\hat{\mathbf{D}}\mathbf{v}) &= \mathbf{0}
\end{align}
if $\hat{\mathbf{D}}^\intercal$ has full rank this implies that 
\begin{align}
  \hat{\mathbf{D}}\mathbf{v} &= \mathbf{0}
\end{align}
Note that $\hat{\mathbf{D}}$ is identical to $\mathbf{D}$ in \eqref{eq:velocity_averaging} up to normalization of each velocity $\mathbf{u}_i^\ell$ separately. This normalization strategy can be seamlessly integrated into the computation of $\mathbf{D}$. Moreover, computing $\mathbf{D}$ is much simpler than computing $\mathbf{F}$.


\subsection{Canonicalization of the Manifold}
\label{sec:app:manifold}

\input{figures/fig_manifold_visualization}

The incidence relation in~\eqref{eq:incidence_final} yields a simple way to visualize the manifold in its canonical form, and also shows the dependence on the line velocity parameters $u^{\ell}_{y}$ and $u^{\ell}_{z}$. To reach this canonical form, we simply rotate the bearing vectors of all events into the line-dependent coordinate frame by replacing $\mathbf{f}' = \mathbf{R}_{\ell} \hat{\mathbf{f}}'$. We visualize this transformation in~\figref{fig:eventail_interpr}, where we transition from raw events in normalized coordinates (A), derotated events (B), and then events in the line reference frame (C). This coordinate frame corresponds with that of a line that is parallel to the camera's $x$-axis. Doing this replacement yields 
\begin{align}
  t' \hat{\mathbf{f}_{j}'}^\intercal (u^{\ell}_{z} \mathbf{e}_{y} - u^{\ell}_{y} \mathbf{e}_{z}) + {\hat{\mathbf{f}_{j}'}}^\intercal \mathbf{e}_{y} = 0 \,,
\end{align}
where $\mathbf{e}_{x/y/z}$ corresponds to the unit vectors in the camera coordinate frame. Distributing and diving out the third component of $\hat{\mathbf{f}_{j}'}$, \ie~transitioning to normalized coordinates in the new line reference frame, we reach
\begin{align}
  0 &= t' (u^{\ell}_{z} \hat{y}^{\ell} - u^{\ell}_{y}) + \hat{y}^{\ell}
  \implies & \hat{y}^{\ell} = \frac{u^{\ell}_{y}t'}{1 + u^{\ell}_{z}t'}
\end{align}
where $\hat{y}^\ell$ is the event $y$-coordinate in normalized, line coordinates. This form describes the shape of the manifold in two dimensions and is visualized in~\figref{fig:eventail_interpr}(ii) for varying $u^{\ell}_{y}$ and $u^{\ell}_{z}$. 

From these visualization we make a number of observations: First, configurations with $u^{\ell}_{z} = 0$ trace straight lines, corresponding to planar manifolds in the line coordinate frame. Note, however, that in the derotated frame (B) these may still be non-planar. Second, we see that $u^{\ell}_{z} < 0$ induces a curvature in the manifold which increases as time progresses. This configuration corresponds to a camera approaching the line, and thus the reduced distance increases the apparent motion, which results in a larger slope. Finally, $u^{\ell}_{z} > 0$ results in flattened curves. This corresponds to cameras retracting from the line, which reduces the apparent motion, and thus reduces the slope in the manifold.

%% file: figures/fig_sim_event_number_analysis.tex
\begin{figure}[t]
  \centering
  \includegraphics[width=0.95\linewidth]{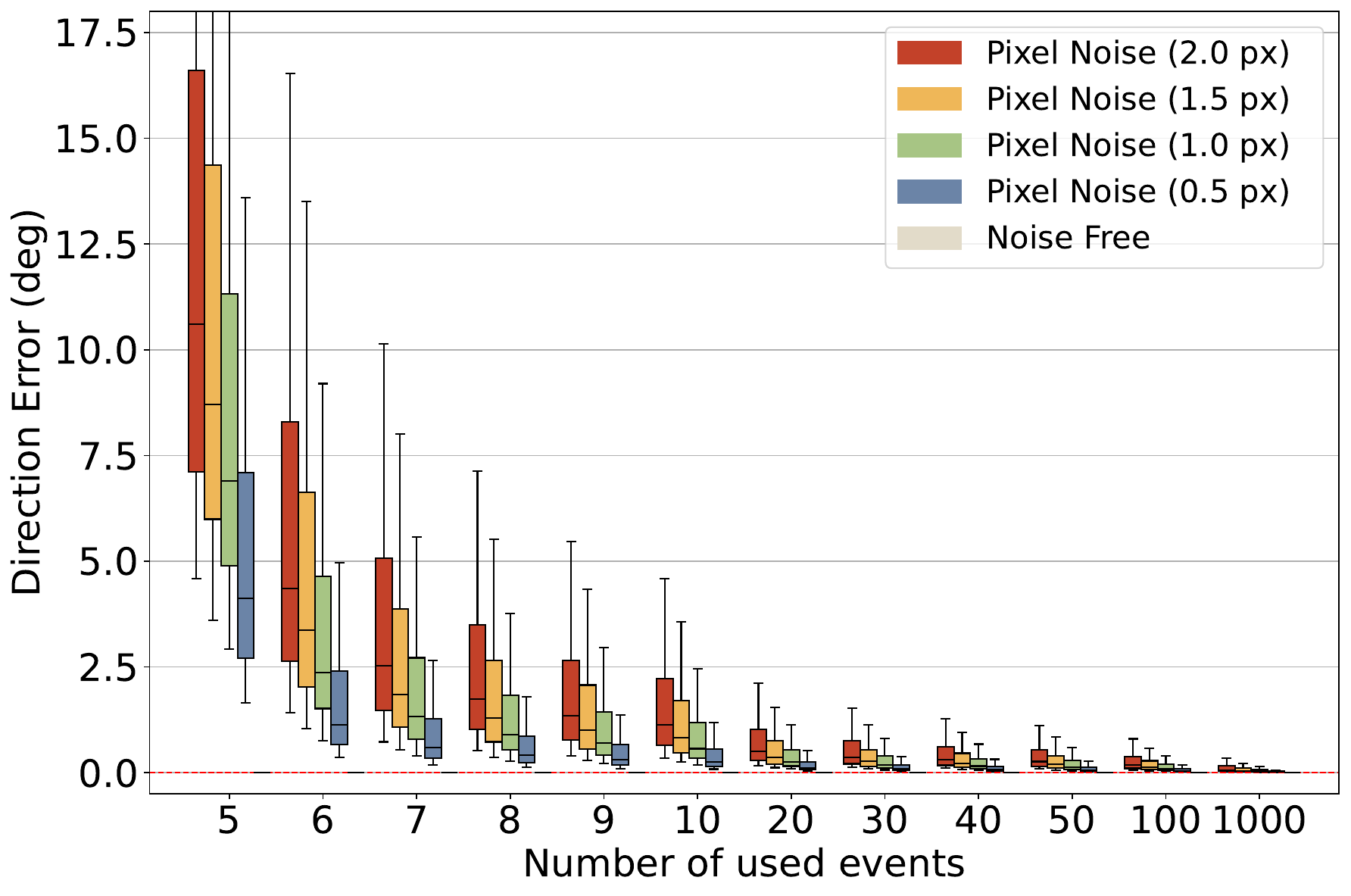} \\ 
  \includegraphics[width=0.95\linewidth]{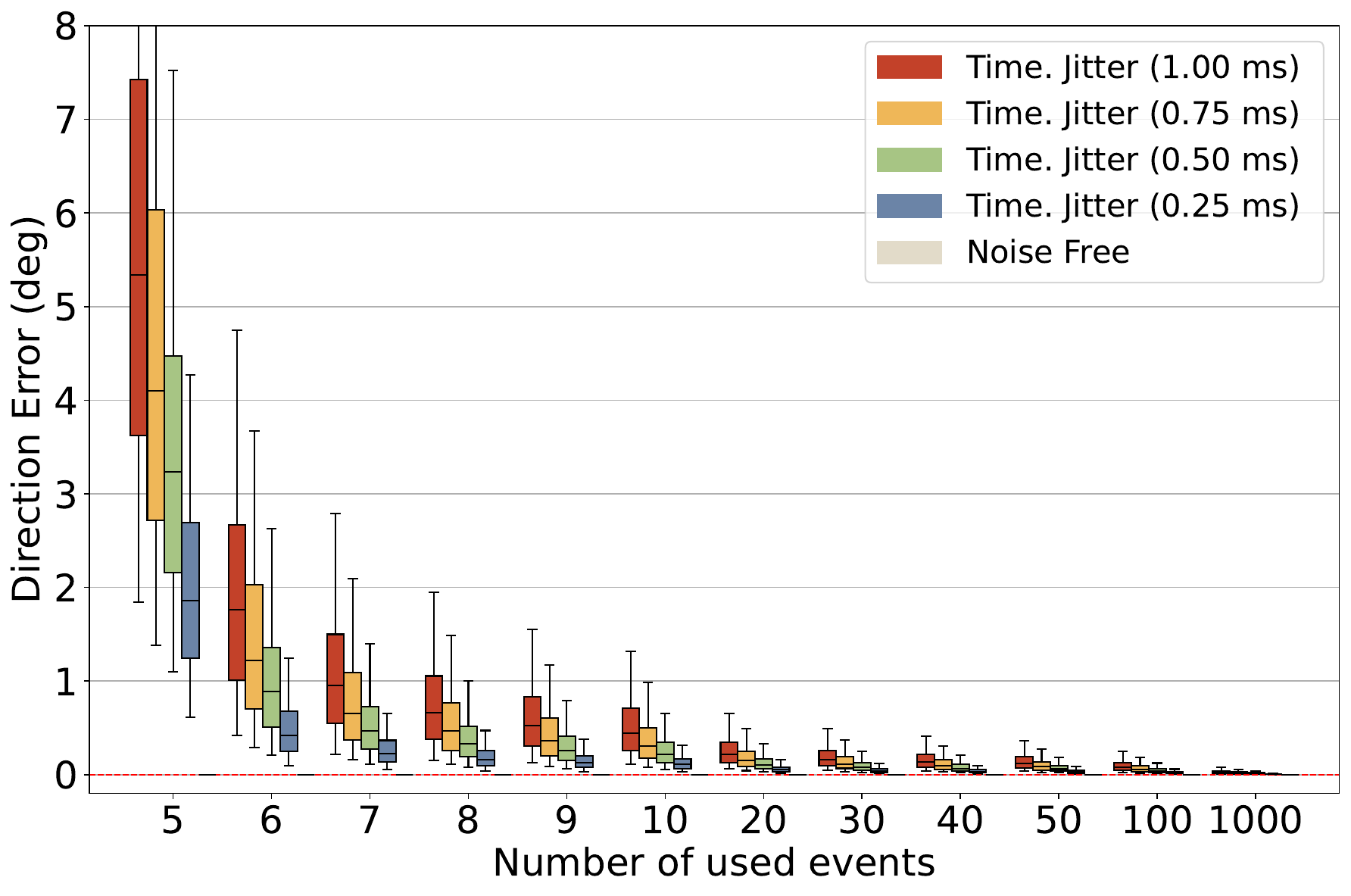} \\
  \includegraphics[width=0.95\linewidth]{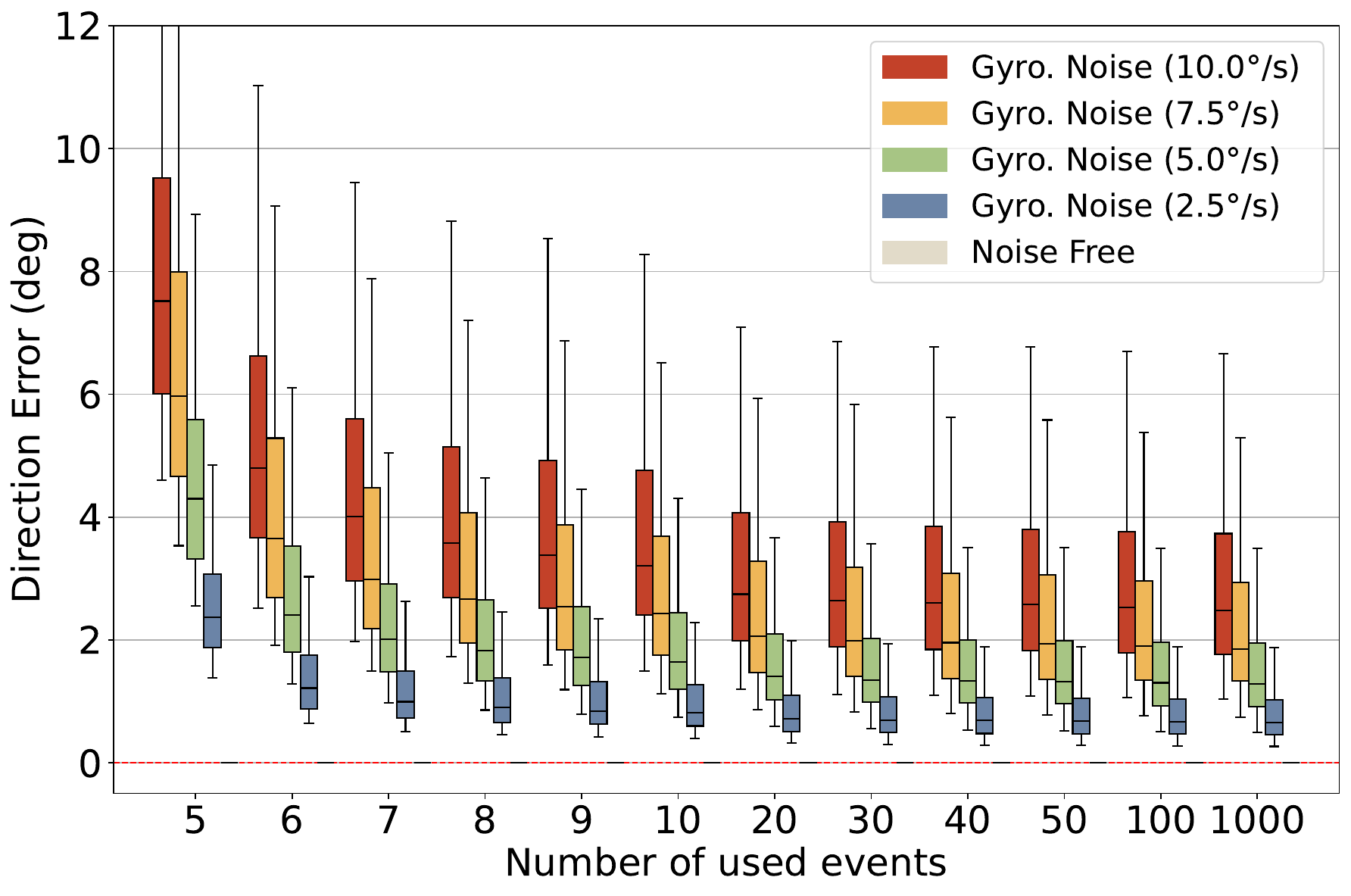}
  \caption{Analysis of the number of used events over three types of representative noise, \ie~pixel noise, timestamp jitter, and gyroscope noise.}
  \label{fig:sim_event_number_analysis}
\end{figure}

%% file: figures/fig_sim_multiple_clusters_analysis.tex
\begin{figure}[t]
  \centering
  \includegraphics[width=0.95\linewidth]{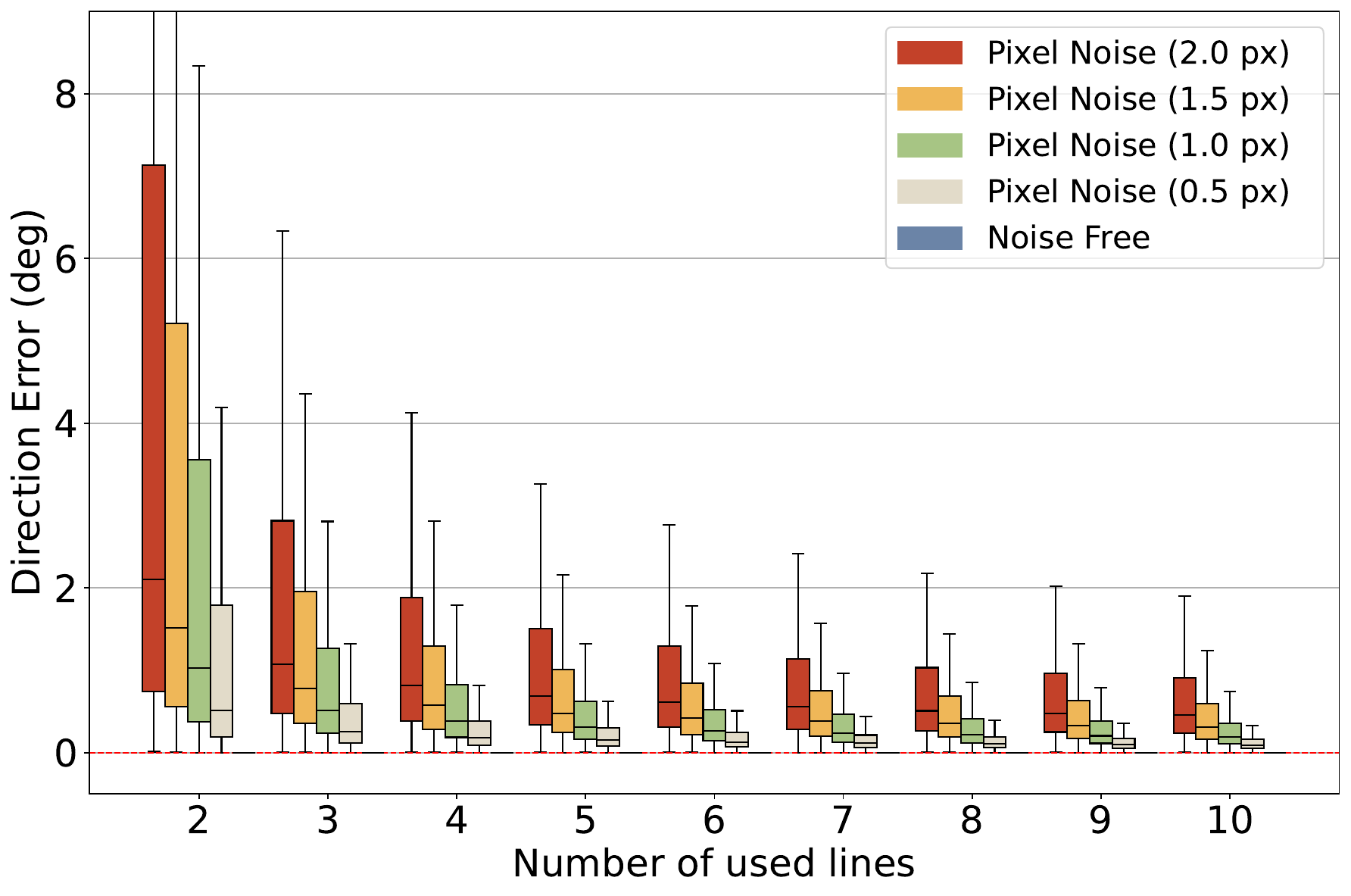} \\
  \includegraphics[width=0.95\linewidth]{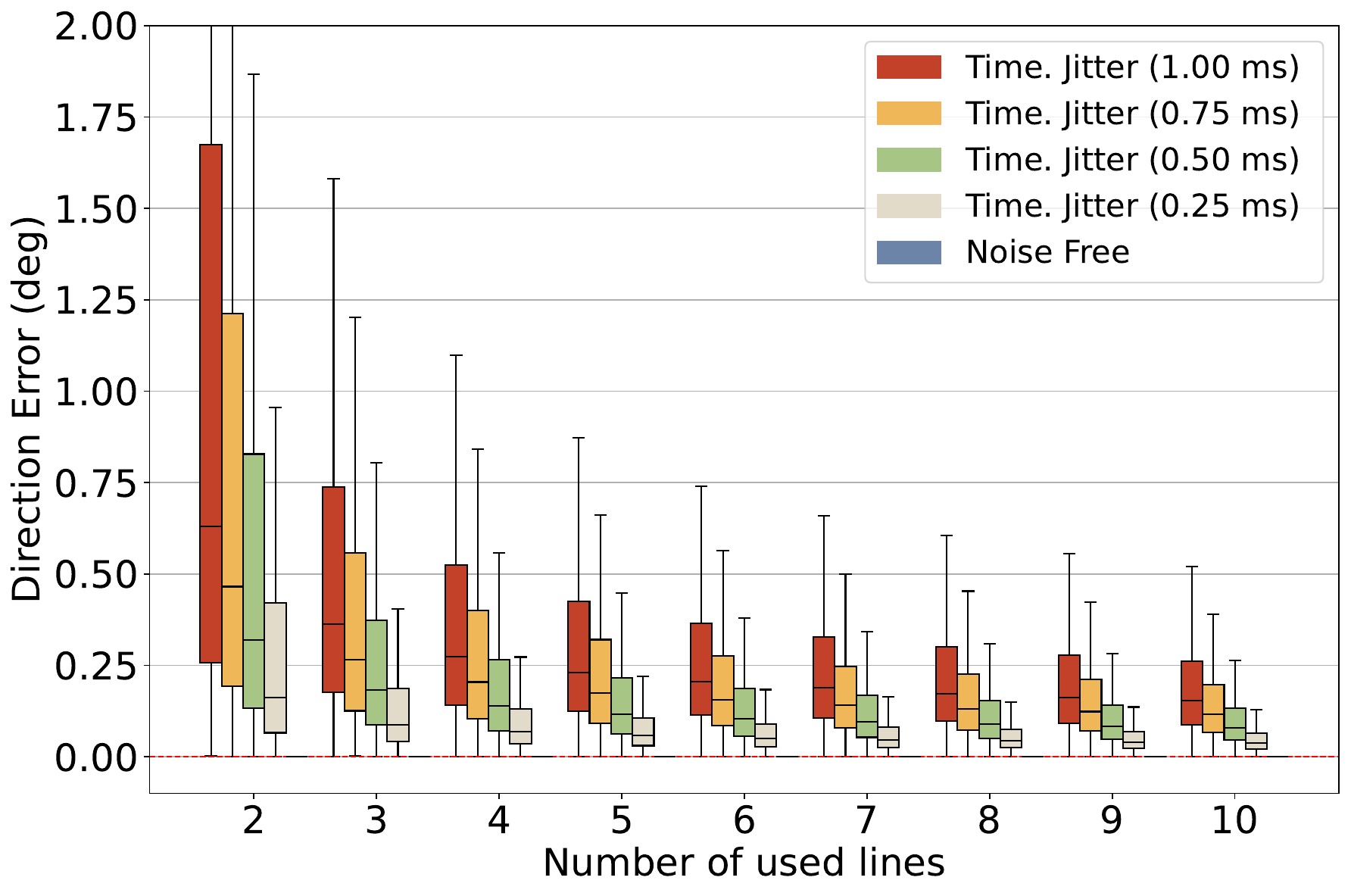} \\
  \includegraphics[width=0.95\linewidth]{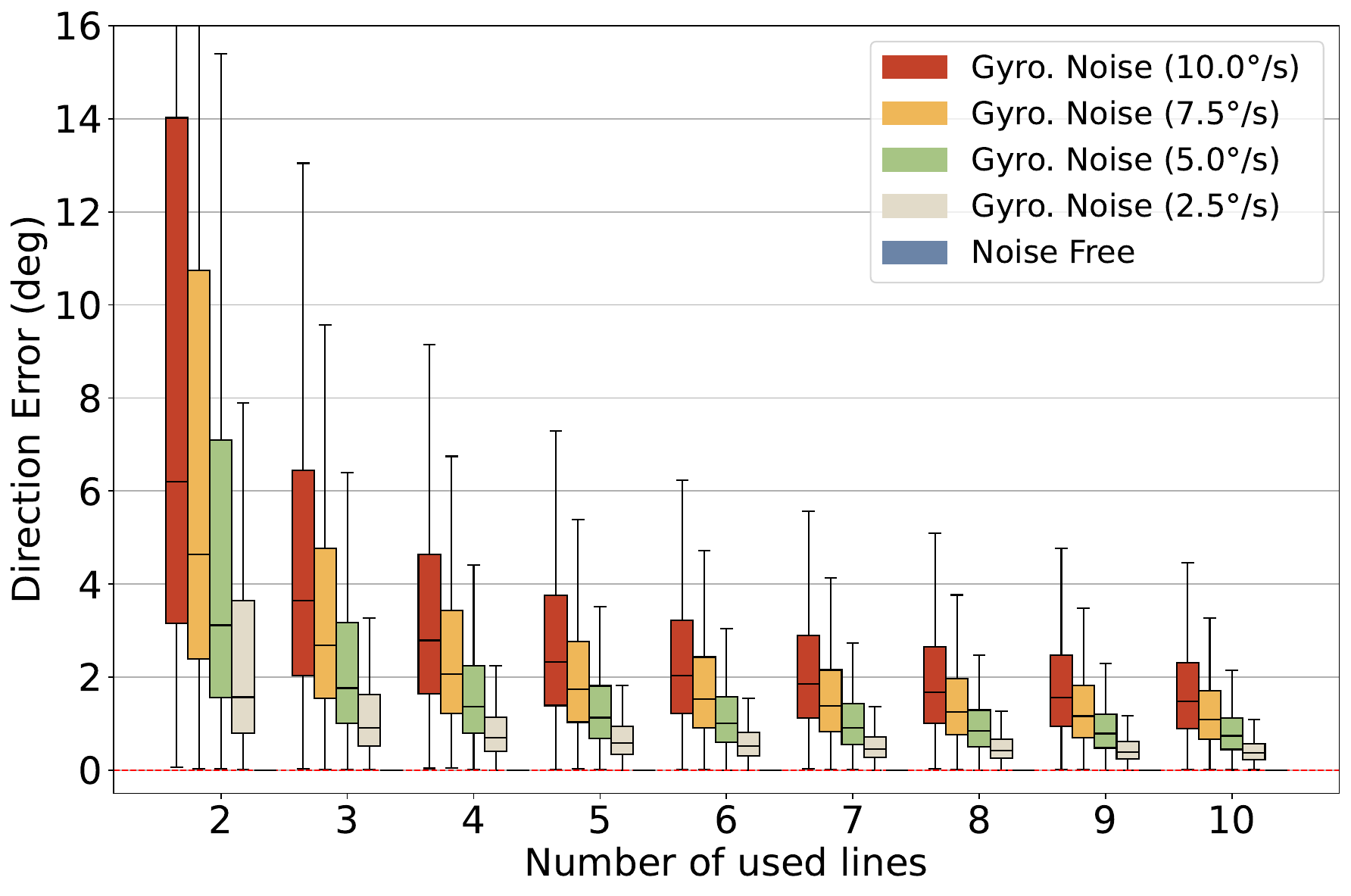}
  \caption{Analysis of the number of used lines over three types of representative noise, \ie~pixel noise, timestamp jitter, and gyroscope noise.}
  \label{fig:sim_multiple_clusters_analysis}
\end{figure}

%% file: figures/fig_manifold_visualization.tex
\begin{figure}[t]
  \centering
  \resizebox{\linewidth}{!}{
  \begin{tabular}{c}
    \includegraphics[height=4cm]{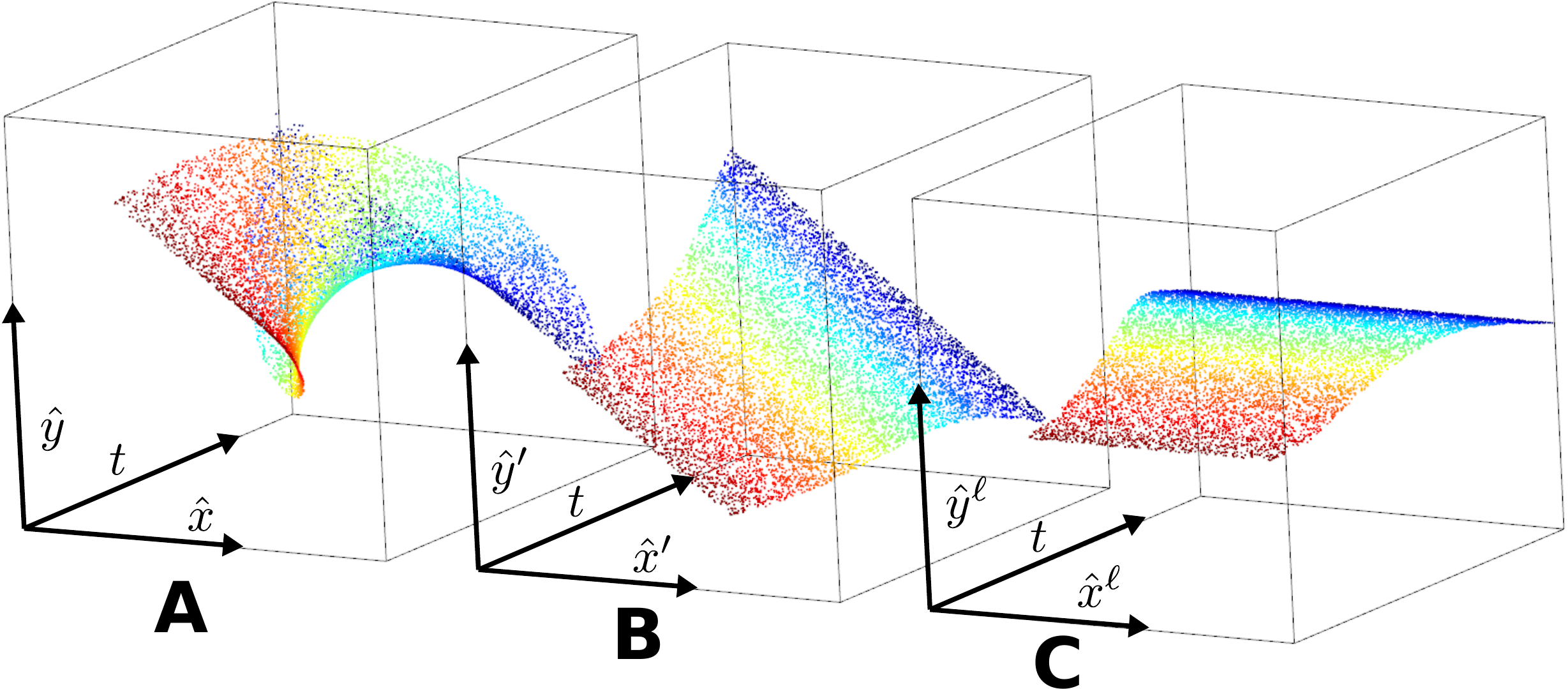} \\
    (i) event transformations \\[1em]
    \includegraphics[width=1\linewidth]{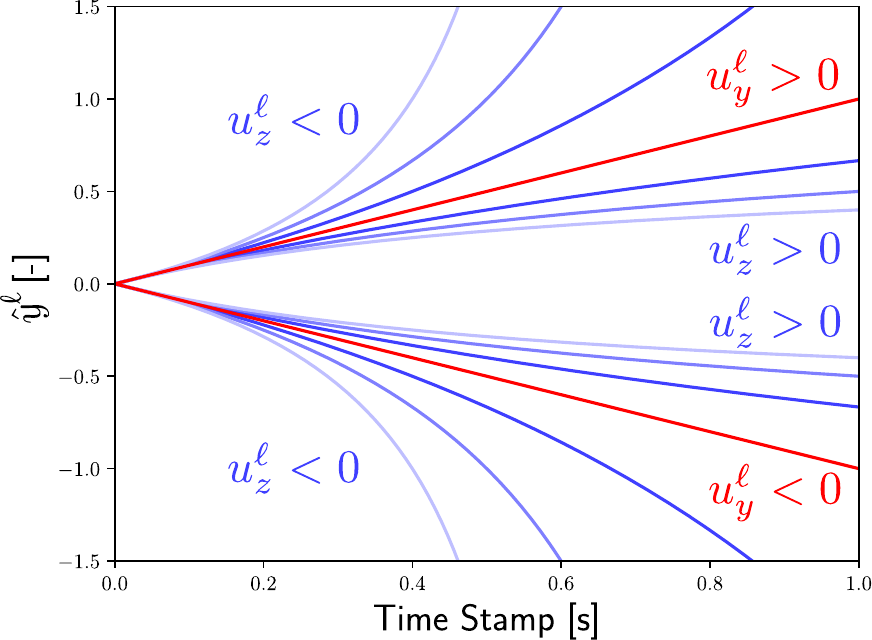} \\
    (ii) eventail in 2D
  \end{tabular}}
  \caption{Events triggered by a line observed by an event camera span a non-linear manifold called \emph{\'eventail} (A). We show that this manifold imposes a linear constraint on the partial camera velocity and line parameters. With this insight, we can design a linear solver for these quantities that is both fast and highly interpretable, and characterize all manifolds (A) by transforming them into canonical form via rotation compensation (B), and transformation into the line coordinate frame (C). In the $\hat{y}^{\ell}t$-plane, these manifolds trace a family of curves (ii), depending on the configuration of $u^{\ell}_{z}$ and $u^{\ell}_{y}$.}
  \label{fig:eventail_interpr}
\end{figure}